%% file: main.tex
\documentclass{article}

\usepackage{microtype}
\usepackage{graphicx}
\usepackage{subfigure}
\usepackage{booktabs} 
\usepackage{verbatim}

\usepackage{hyperref}

\usepackage[accepted]{icml2021}

\usepackage{amssymb, amsmath, mathtools}
\usepackage{adjustbox} 
\DeclareMathOperator*{\argmax}{arg\,max}
\DeclareMathOperator*{\argmin}{arg\,min}
\DeclareMathOperator{\E}{\mathop{\mathbb{E}}}

\usepackage{pythonhighlight}
\usepackage{listings}

\definecolor{gray}{gray}{0.5}
\colorlet{commentcolour}{green!50!black}

\colorlet{stringcolour}{red!60!black}
\colorlet{keywordcolour}{magenta!90!black}
\colorlet{exceptioncolour}{yellow!50!red}
\colorlet{commandcolour}{blue!60!black}
\colorlet{numpycolour}{blue!60!green}
\colorlet{literatecolour}{magenta!90!black}
\colorlet{promptcolour}{green!50!black}
\colorlet{specmethodcolour}{violet}

\lstdefinestyle{mypython}{
language=python,
showtabs=true,
tab=,
tabsize=2,
basicstyle=\ttfamily\scriptsize,
stringstyle=\color{stringcolour},
showstringspaces=false,
alsoletter={1234567890},
otherkeywords={\%, \}, \{, \&, \|},
keywordstyle=\color{keywordcolour}\bfseries,
emph={and,break,class,continue,def,yield,del,elif ,else,%
except,exec,finally,for,from,global,if,import,in,%
lambda,not,or,pass,print,raise,return,try,while,assert,with},
emphstyle=\color{blue}\bfseries,
emph={[2]True, False, None},
emphstyle=[2]\color{keywordcolour},
emph={[3]object,type,isinstance,copy,deepcopy,zip,enumerate,reversed,list,set,len,dict,tuple,xrange,append,execfile,real,imag,reduce,str,repr},
emphstyle=[3]\color{commandcolour},
emph={Exception,NameError,IndexError,SyntaxError,TypeError,ValueError,OverflowError,ZeroDivisionError},
emphstyle=\color{exceptioncolour}\bfseries,
morecomment=[s]{"""}{"""},
commentstyle=\color{commentcolour}\slshape,
emph={[4]ode, fsolve, sqrt, exp, sin, cos,arctan, arctan2, arccos, pi,  array, norm, solve, dot, arange, isscalar, max, sum, flatten, shape, reshape, find, any, all, abs, plot, linspace, legend, quad, polyval,polyfit, hstack, concatenate,vstack,column_stack,empty,zeros,ones,rand,vander,grid,pcolor,eig,eigs,eigvals,svd,qr,tan,det,logspace,roll,min,mean,cumsum,cumprod,diff,vectorize,lstsq,cla,eye,xlabel,ylabel,squeeze},
emphstyle=[4]\color{numpycolour},
emph={[5]__init__,__add__,__mul__,__div__,__sub__,__call__,__getitem__,__setitem__,__eq__,__ne__,__nonzero__,__rmul__,__radd__,__repr__,__str__,__get__,__truediv__,__pow__,__name__,__future__,__all__},
emphstyle=[5]\color{specmethodcolour},
emph={[6]assert,yield},
emphstyle=[6]\color{keywordcolour}\bfseries,
emph={[7]range},
emphstyle={[7]\color{keywordcolour}\bfseries},
literate=*%
{:}{{\literatecolour:}}{1}%
{=}{{\literatecolour=}}{1}%
{-}{{\literatecolour-}}{1}%
{+}{{\literatecolour+}}{1}%
{*}{{\literatecolour*}}{1}%
{**}{{\literatecolour{**}}}2%
{/}{{\literatecolour/}}{1}%
{//}{{\literatecolour{//}}}2%
{!}{{\literatecolour!}}{1}%
{[}{{\literatecolour[}}{1}%
{]}{{\literatecolour]}}{1}%
{<}{{\literatecolour<}}{1}%
{>}{{\literatecolour>}}{1}%
{>>>}{\pythonprompt}{3}%
,%
frame=trbl,
rulecolor=\color{black!40},
backgroundcolor=\color{white},
breakindent=.5\textwidth,frame=single,breaklines=true%
}

\icmltitlerunning{Learning Routines for Effective Off-Policy Reinforcement Learning}

\begin{document}

\twocolumn[
\icmltitle{Learning Routines for Effective Off-Policy Reinforcement Learning}
\icmlsetsymbol{equal}{*}

\begin{icmlauthorlist}
\icmlauthor{Edoardo Cetin}{kcl}
\icmlauthor{Oya Celiktutan}{kcl}
\end{icmlauthorlist}

\icmlaffiliation{kcl}{Centre for Robotics Research, Department of Engineering, King's College London}

\icmlcorrespondingauthor{Edoardo Cetin}{edoardo.cetin@kcl.ac.uk}

\icmlkeywords{Machine Learning, ICML}

\vskip 0.3in
]

\printAffiliationsAndNotice{}

\begin{abstract}
The performance of reinforcement learning depends upon designing an appropriate action space, where the effect of each action is measurable, yet, granular enough to permit flexible behavior. So far, this process involved non-trivial user choices in terms of the available actions and their execution frequency. We propose a novel framework for reinforcement learning that effectively lifts such constraints. Within our framework, agents learn effective behavior over a \textit{routine space}: a new, higher-level action space, where each \textit{routine} represents a set of `equivalent' sequences of granular actions with arbitrary length. Our routine space is learned end-to-end to facilitate the accomplishment of underlying off-policy reinforcement learning objectives. We apply our framework to two state-of-the-art off-policy algorithms and show that the resulting agents obtain relevant performance improvements while requiring fewer interactions with the environment per episode, improving computational efficiency.
\end{abstract}

\input{sections/1Intro}
\input{sections/2RelatedWork}
\input{sections/3Preliminaries}

\input{sections/4Method}

\input{sections/5Algorithm}

\input{sections/6Experiments}

\input{sections/7Conclusion}

\section*{Acknowledgements}
Edoardo Cetin would like to acknowledge the support from the Engineering and Physical Sciences Research Council [EP/R513064/1]. Additionally, Oya Celiktutan would like to acknowledge the support from the LISI Project, funded by the Engineering and Physical Sciences Research Council [EP/V010875/1].

\bibliography{bibliography}
\bibliographystyle{icml2021}

\newpage

\appendix
\input{appsections/APseudocode}
\input{appsections/BIntegration}
\input{appsections/CHyper}

\input{appsections/DFigarTD3}

\input{appsections/DFullRes}
\input{appsections/FRoutineSpaceAnalysis}
\end{document}

%% file: sections/1Intro.tex
\section{Introduction}

The applicability of machine learning has seen tremendous advancements with the advent of deep, expressive models \citep{lecun-deep}. These models enabled to tackle complex problems given large quantities of data through end-to-end training, removing bottlenecks related to hand-designed features. Such trend has seen success also in reinforcement learning, where deep learning enabled the achievement of significant milestones \citep{dqn, alphazero} and the specification of practical algorithms \citep{ddpg, td3, sac}.

However, within the reinforcement learning framework, there are still many fixed components, related to the agent's interface with the environment, that are the result of expert-engineering and are often quite influential on the performance \cite{rl-real-world-design}. Algorithms that learn also these additional components end-to-end would alleviate some of the deployment burdens and likely improve final performance given enough data. One of such components is the agent's action space. The performance of current reinforcement learning algorithms is dependent on reasoning with an expressive set of actions that have a tangible effect on the current state. Specifying this generally includes setting an environment-specific fixed frequency with which the agent alternates reasoning and executing behavior. 

On the other hand, humans are capable of effectively performing actions to the lowest level of granularity while only directly reasoning with higher-level abstractions of what they \textit{intend} to do. 
This can be attributed to their observable ability of acquiring, through repetition of a particular task, temporally-extended motor-skills that can be re-enacted without conscious thought \cite{human-motor-learning}. 
Doing so allows to abstract away the details regarding the composition of each motor-skill, thus, minimizing the required reasoning to solve the task. We will refer to these motor-skills, acquired with the objective of solving a particular task more efficiently as \textit{routines}.

From these observations, we design a new class of reinforcement learning algorithms that learn behavior together with a new `action space', 
 which we name the \textit{routine space}. In this space, each \textit{learned routine} represents higher-level concepts of what the agent intends to achieve with respect to the task goal and can be mapped to a whole 
 set of `equivalent' variable-length sequences of action primitives. Within our framework, this space is learned end-to-end to facilitate the underlying objectives of any base reinforcement learning algorithm. 

Our experimental results demonstrate that utilizing our proposed \textit{routine framework} improves the performance of two different off-policy reinforcement learning algorithms tested on the environments from the DeepMind Control Suite \cite{dmc}. Moreover, using our framework, agents need to reason only after experiencing the outcome of each routine rather than each action. Therefore, they are able to query their policy much more infrequently by learning to perform longer routines from states that do not require a fine level of control. Practically, this enables for computationally efficient deployment, faster data-collection, and easier real-time inference \cite{challenges-rl}. For access to our open-source implementations, please visit \texttt{sites.google.com/view/routines-rl/}.

In summary, the contribution of this work is three-fold:
\begin{itemize}
    \item We introduce the concept of routines in reinforcement learning and discuss its inherent advantages.
    \item We propose effective methods to integrate the routine framework with off-policy reinforcement learning and describe two new algorithms, based on the \textit{Twin Delayed DDPG} \cite{td3} and \textit{Soft Actor-Critic} \cite{sac-alg} algorithms. 
    \item We provide an extensive evaluation on the DeepMind Control Suite and show that the routine framework boosts performance and computational efficiency, with agents requiring substantially fewer policy queries to successfully act during an episode.
\end{itemize}

%% file: sections/2RelatedWork.tex
\section{Related Work}

Learning higher-level abstractions of an agent's actuation interface has been a long-studied problem in reinforcement learning under several similar conceptual frameworks. \textit{Options} \citep{opt-intro1, opt-intro2} represent one of such frameworks in which an agent tries to learn a set of sub-policies, where execution is alternated according to relative termination conditions and a higher-level controller. Learning these sub-policies has been attempted by specifying subgoals \citep{opt-sub-1, opt-sub-2} and, more recently, end-to-end optimization \citep{option-critic} with varying degrees of success. Yet, it still has remained an open problem. The routine framework could be viewed as an unconventional instance of the options framework, where the routine space could represent a whole space of task-dependent `sub-policies' corresponding to fixed distributions over action sequences. 
Similarly, several works in hierarchical reinforcement learning have attempted to partition an agent's policy into low-level and high-level controllers. 
One of the key challenges, though, has remained recovering effective low-level behavior.
To tackle this challenge, some of these works considered feeding extrinsic rewards for achieving heuristically useful behavior \citep{hierarchical-subgoals-small-prior} or solving a diverse range of subtasks \citep{hierarchical-subgoals-mine, hierarchical-subgoals-robot}. Alternatively, other works studied intrinsic objectives related to exploration measures \citep{hierarchical-subgoals-intrinsic-1, hierarchical-subgoals-intrinsic-2}. Recently, \citet{NDP} also explored hierarchical structures, utilizing a dynamical system to achieve some higher-level reparameterization of the action space, incorporating effective inductive bias into the agent's model.

\textit{Macro-actions} \cite{macro-intro} represent another concept related to routines and options, proposing to use particular fixed sequences of actions to build a higher-level abstraction of the action space. 
A simple way of incorporating macro-actions into the learning framework is to specify policies reasoning with action repetitions. Several works showed that such practice can speed up learning and facilitate exploration \citep{actreps_mot0, actreps_mot1, actreps_mot2}. Particularly, \citet{action-reps-discrete} proposed a manually augmented discrete action space where each action would be present at two different temporal resolutions, while \citet{action-reps-policy-factor} and \citet{act-rep-cr} factorized the policy to output both an action and the corresponding number of repetitions.
Closer to our work, the framework from \citet{macro-STRAWS} utilized more expressive macro-actions by keeping a running plan of future behavior to execute, adaptively updated with attentive reading and writing operations. 

In line with our conceptualization, several works have considered building explicit embedding spaces of behavior to directly facilitate learning. Within this area, some works considered learning representations of individual actions based on the environment dynamics, in order to aid exploration in large discrete action spaces \citep{single-action-rep-discrete-1, single-action-rep-discrete-2}. Other works also considered building embeddings of temporally-extended behavior by exploiting demonstrations \citep{demos-act-rep} or representations of target `goal-states' \citep{FuN, HIRO}. Sharing some commonalities with our method, \textit{SeCTAR} \citep{traj-embeddings-auto-encoder} used a model to `auto-encode' fixed-length sequences of states, and learned a low-level policy to follow the reconstructed state trajectory conditioned on the lower-dimensional embeddings. Additionally, \citet{actseq-rep-dynamics-aware-emb} proposed to learn representations of fixed-length action sequences to be maximally useful for predicting state transitions. While many of these representations demonstrated meaningful results improving sample-efficiency and performance, unlike routines, they still required specifying explicit heuristic objectives to obtain an effective encoding.

%% file: sections/3Preliminaries.tex
\section{Preliminaries}

\subsection{Markov Decision Process}

In reinforcement learning, the agent's attempted task can be described as a Markov Decision Process (MDP), defined as $(S, A, r, P, p_0, \gamma)$. At each time-step, the agent experiences a state in its state space, $s \in S$, and performs an action in its action space, $a \in A$. $P: S \times A \times S \rightarrow \mathbb{R}$ and $p_0: S\rightarrow \mathbb{R}$ represent the transition dynamics and the initial state distribution, determining the likelihoods of experiencing particular transitions. $r: S \times A \rightarrow \mathbb{R}$ is the reward function, outputting numerical measures representing the usefulness of observed behavior towards achieving the task's objective. The reinforcement learning goal is to obtain a policy $\pi$ to maximize the expected overall performance quantified by the sum of discounted experienced rewards:
\begin{equation}
\label{pi_obj}
\argmax_\pi\E_{p_\pi(\tau)}\left[ \sum^{\infty}_{t = 0}\gamma^t r (s_t, a_t) \right],
\end{equation}
where $p_\pi(\tau)$ represents the distribution of trajectories $\tau$ encountered by the agent from its interaction with the environment. 

\subsection{Off-Policy Learning}

\label{offpol}

At any point during the agent's trajectory, we can calculate its expected future performance after taking a particular action, quantified by the expected discounted sum of future rewards. This quantity is called the Q-function:
\begin{equation}
\label{q_fn}
Q^\pi(s, a) = \E_{p_\pi(\tau|s_0{=}s, a_0{=}a)}\left[ \sum^{\infty}_{t = 0}\gamma^t r (s_t, a_t)\right].
\end{equation}
The Q-function can be represented by a parameterized model $Q_\phi$ and learned for any policy $\pi$ by iteratively minimizing a squared temporal difference (TD-) loss of the form $\argmin_\phi \E_{(s, a, s', r) \sim B}\left[(Q^\pi_\phi(s, a) - y)^2\right]$, where the TD-targets $y$ are obtained by computing the Bellman backups $y = r + \gamma Q^\pi_{\phi'}(s', \pi(s'))$. This objective is often optimized by sampling batches of uncorrelated transitions from a replay buffer $B$ and using a slowly updated \textit{target} network $Q^\pi_{\phi'}$ to compute the TD-targets $y$.

The agent's policy can also be represented by a parameterized model $\pi_\theta$ and iteratively improved by maximizing its relative Q-function, following the policy gradient theorem \citep{pg-thm, dpg}:
\begin{equation}
\label{pi_pg_obj}
\argmax_\theta\E_{s\sim B, a\sim \pi_\theta(\cdot|s)}\left[Q_\phi^{\pi_\theta} (s,a)\right].
\end{equation}

\subsection{Advanced Policy Gradient Algorithms}
\label{advanced-pga}
Current state-of-the-art off-policy algorithms build upon the concepts expressed in Section~\ref{offpol} and propose further complementary ideas. In particular, we will be considering the \textit{Twin-delayed DDPG} (\textit{TD3}) algorithm \cite{td3}, which proposes several practical innovations to aid the optimization of the policy gradient objective such as learning independently two Q-networks, $Q^\pi_{\phi_1}, Q^\pi_{\phi_2}$, and updating the policy less frequently for stability. We will also be considering the \textit{Soft Actor-Critic} (\textit{SAC}) algorithm \cite{sac, sac-alg}, which proposes to optimize an augmented maximum-entropy reinforcement learning objective \cite{maxentobj}:
\begin{equation}
\label{pi_obj_ent}
\argmax_\pi\E_{p_\pi(\tau)}\left[ \sum^{\infty}_{t = 0}\gamma^t r (s_t, a_t) + \alpha H(\pi(a_t|s_t))\right].
\end{equation}

%% file: sections/4Method.tex
\section{The Routine Framework}
\label{sec:method}
\subsection{Routines Formalization}

For continuous control problems, each action can be represented by a vector in $\mathbb{R}^a$ corresponding to some actuation inputs, e.g, the torque to be applied to each of an agent's joints. Generally, the execution length of each action is an environment-specific fixed hyper-parameter, determining with which frequency the agent reasons and interacts with the environment. As such, this value can often influence the resulting agent's performance. 

We represent each routine, $n \in N$, as a vector in $\mathbb{R}^n$, corresponding to the \textit{routine space}. Each routine is tied with a subset of $S^A$:
\begin{equation}
    S^A=\left\{(a_1, a_2, ..., a_l):l\leq L\right\},
\end{equation}
corresponding to the set of all action sequences up to some maximum length $L$, built from the original action space $A$. Such maximum length limitation ensures that each routine has a finite result when actuated. Conceptually, all the action sequences tied with a given routine should accomplish related results with respect to the task's objective, giving rise to a many-to-one mapping. Unlike actions the meaning of each routine should not depend on a specific agent's actuators and can even change during optimization. 

Within our framework, the policy $\pi$ represents a conditional distribution over routines rather than actions. Additionally, we define a routine Q-function, defined on the set of all states and routines that represents the expected future performance when starting a trajectory by actuating a particular routine: 
\begin{equation}
\label{q_fn}
Q^\pi(s, n) = \E_{p_\pi(\tau|s_0{=}s, n_0{=}n)}\left[ \sum^{\infty}_{t = 0}\gamma^t r (s_t, a_t) \right].
\end{equation}

\subsection{Learning Routines}
\label{learning-routines}

\textbf{Routine decoder}. In order to execute any given routine in the environment, we define a model that specifies a mapping from routines to action sequences in $S^A$. 
We call this model the \textit{routine decoder}, $D_{\omega_1}: N \mapsto (\mathbb{R}^{a \times L}, \mathbb{R}^{L-1})$. In its simplest form, we represent this model with a neural network $D_{\omega_1}$, taking as input a routine $n$, and outputting an action sequence of length $L$, $a_{1:L}^n$, together with a list of early termination probabilities $e^n_{1:L-1}$. The early termination probabilities can then be used to sample an action sequence in $S^A$, by representing the likelihood of `cutting-off' the outputted action sequence $a_{1:L}^n$ into a subsequence $a_{1:l}^n$ for different lengths $l$, as described by the following equations:
\begin{align}
    p(a^n_{1:l}|n) = e^n_l &\prod_{j=1}^{l-1} (1-e^n_j) \text{ for }  1\leq l<L, \nonumber \\
    p(a^n_{1:L}|n) = &\prod_{j=1}^{L-1} (1-e^n_j).
    \label{seg-len-prob}
\end{align}
Therefore, to query the agent's behavior from state $s$, we evaluate the agent's policy $\pi$ to obtain a routine $n$. Then, we feed $n$ to the routine decoder and utilize its outputs to sample the routine length $l$ with its corresponding action sequence $a^n_{1:l}$ to execute. 

\textbf{Routine encoder}. Since the meaning of each routine can evolve during training, 
recording the representations of performed routines within the replay buffer would not be useful. Instead, we record the resulting executed action sequences, which have an invaried effect on the environment. Thus, to use replay buffer data for learning the routine Q-function, we require a model to translate back action sequences from the replay buffer to their corresponding routines. We call this model the \textit{routine encoder}, $E_{\omega_2}: S^A \mapsto N$. We represent this model as an additional neural network $E_{\omega_2}$ taking as input action sequences $a_{1:l}$ and outputting the corresponding routine $n^{a_{1:l}}$. 

\begin{figure}
  \centering
  \includegraphics[width=0.9\columnwidth]{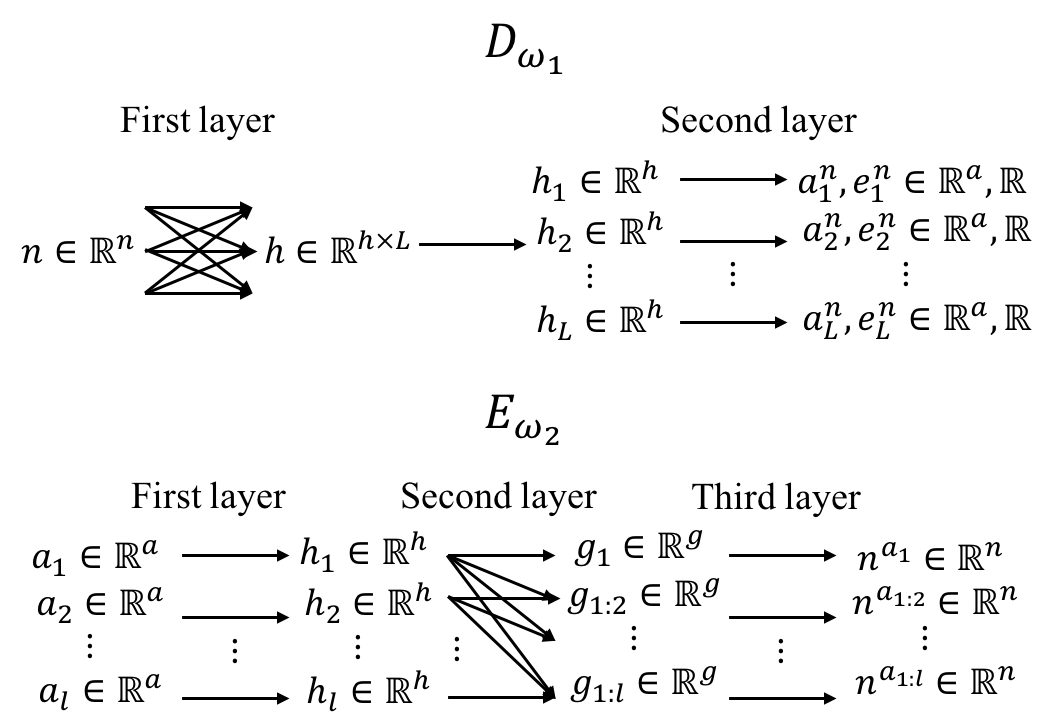}
  \caption{Simplified network structures of the routine decoder $D_{\omega_1}$ (Top) and encoder $E_{\omega_2}$ (Bottom).}
  \label{figure:enc-dec-models}
\end{figure}

\textbf{Models parameterization.} The routine decoder is parameterized as a 2-layer neural network, taking as input a routine $n$. Particularly, the first layer of the decoder produces a hidden representation that is further split into $L$ separate sub-representations. These sub-representations are then processed independently by the second layer to obtain $L$ corresponding actions $a_{1:L}$ and early termination probabilities $e_{1:L}$. 
The routine encoder is parameterized as a 3-layer neural network, taking as input sequences of actions $a_{1:l}$. The first layer starts by embedding independently the actions into $l$ hidden representations. Subsequently, the second layer further embeds the $l$ produced representations through a set of $L$ `index-specific' weights and ultimately sums them over, obtaining a unique aggregated representation for the whole sequence. Note that by keeping a running sum of the embeddings we are also able to obtain the aggregated representations for all action sub-sequences $\{a_{1:j}: j<l\}$ as byproducts of this computation. The final layer, then, processes these aggregated representations outputting the corresponding routine representation $n^{a_{1}}, n^{a_{1}, a_{2}}, ...,  n^{a_{1:l}}$. We provide a visual representation of these models in Figure~\ref{figure:enc-dec-models}.

\textbf{Consistency objectives.} The routine decoder and encoder should represent a many-to-one mapping where different action sequences with equivalent effect on the environment are mapped to individual routines. To achieve this, we propose to impose two consistency objectives to tie the routine decoder $D_{\omega_1}$ and encoder $E_{\omega_2}$ models. The first objective is to enforce the desired many-to-one consistency, ensuring that reconstructed action sequences are mapped back to the same routine that generated them. Practically, we optimize the routine decoder to decode any particular routine $n$ into an action sequence $a^n_{1:l}$ that would be mapped back to the same routine by the routine encoder $n^{a^n_{1:l}}$. We achieve this by specifying an L2 objective for reconstructing the starting routine representation $J_{mto}(\omega_1)$:

\begin{equation}
\label{E_cons}
\argmin_{\omega_1} \E_{B, \pi} \left[||E_{\omega_2}(D_{\omega_1}(n))) - n||^2\right].
\end{equation} %
The second objective serves to learn the early termination probabilities $e_{1:L-1}$ and is motivated by the prior understanding that, in most tasks, only action sequences with the same length $l$ should be equivalent towards achieving a target goal. 
Particularly, this is enforced by minimizing a binary cross-entropy loss for the early termination probability in a reconstructed action-sequence of length $l$, pushing $e_l$ to 1 and $e_j$ to 0 for all $j<l$: 

\begin{equation}
\label{lc_loss}
L_{lc}(e_{1:L-1}) = \sum_{j=1}^{l-1}\left(log(1-e_j)\right) - log(e_l).
\end{equation} %

This loss is minimized with respect to action sequences sampled from the replay buffer and the resulting objective, $J_{lc}(\omega)$, is optimized for both the routine encoder and routine decoder parameter sets $\omega = \{\omega_1 \cup \omega_2\}$:
\begin{equation}
\label{lc_obj}
\argmin_\omega \E_B\left[L_{lc}(D_{\omega_2}(E_{\omega_1}(a_{1:l})))\right].
\end{equation} %

\subsection{End-to-End Optimization for Off-Policy Learning} 

\label{end-to-end-opt}

\begin{figure}
  \centering
  \includegraphics[width=0.9\columnwidth]{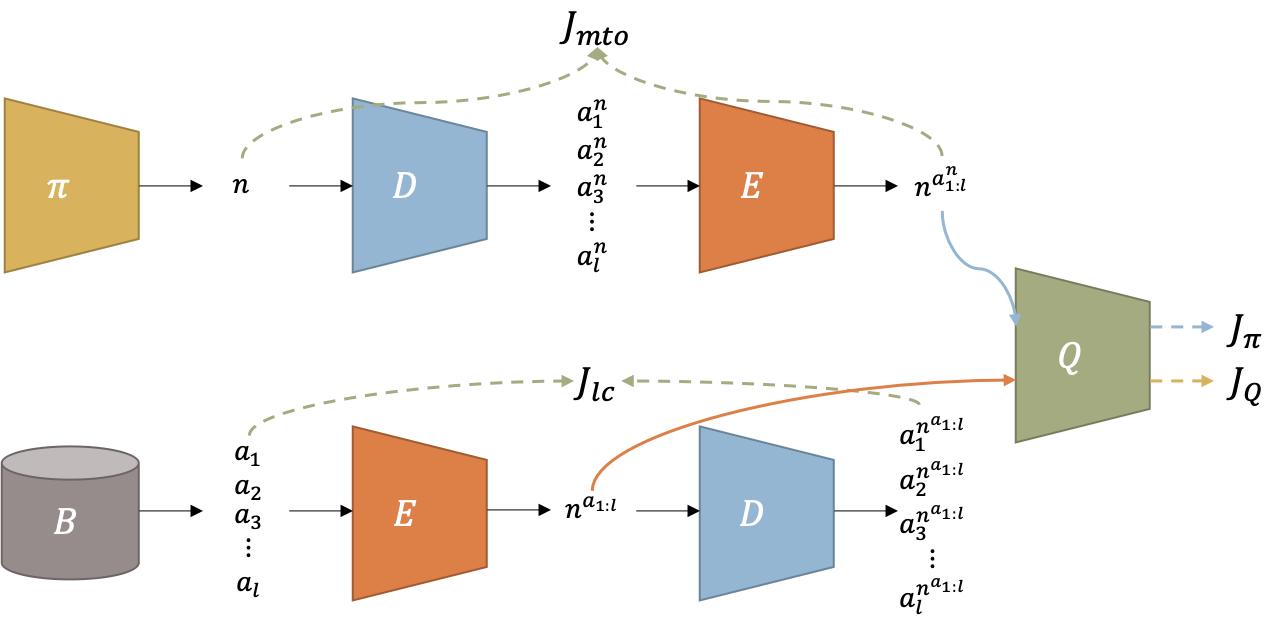}
  \caption{Simplified model optimizations in the routine framework. The top path represents the joint optimization of the policy gradient $J_\pi$ and many-to-one consistency $J_{mto}$ objectives. The bottom path represents the joint optimization of the temporal difference $J_Q$ and length consistency $J_{lc}$ objectives.}
  \label{figure:optimization-flow}
\end{figure}

An \textit{effective} routine space must strive to facilitate the underlying reinforcement learning task. Hence, the \textit{driving} learning signal used to obtain the routine space comes directly from optimizing the objectives of the underlying reinforcement learning algorithms. In this section, we first formulate these objectives in the context of applying the routine framework to off-policy reinforcement learning 
and, subsequently, detail how to perform the end-to-end optimization.

\textbf{Learning the routine Q-function.} To integrate the considered off-policy algorithms with the routine framework, we need to specify a learning procedure to recover an accurate Q-function for every state and routine. We take a \textit{natural} approach to perform this, and optimize a modified squared TD-loss, similarly to how this optimization would be carried out for actions. Particularly, for each sampled state $s$  we collect the sequence of $L$ subsequent actions $a_{1:L}$, next states $s_{1:L}'$ and rewards $r_{1:L}$ from the replay buffer to perform $L$ temporal difference updates in parallel. We encode each `sub-sequence' of consecutive actions, starting from $a_1$, into their corresponding routines $n^{a_{1:l}} = E_{\omega_1}(a_{1:l})$.
This can be performed efficiently with our routine encoder architecture, feeding only the original full-length sequence $a_{1:L}$. The Q-function objective is then to minimize $J_Q$:
\begin{equation}
\label{qi_obj}
\argmin_\phi \E_{B, E_{\omega_2}}\left[\sum_{l=1}^{L}(Q_\phi^\pi(s, n^{a_{1:l}})-y_l)^2\right].
\end{equation}
The TD-targets, $y$, of this optimization are computed as the discounted sum of rewards obtained from executing actions $a_{1:l}$ together with the target Q value at next state $s_{l}'$:
\begin{equation} 
\label{td-targets-int}
y_l = \sum^l_{j=1}\left(\gamma^{(j-1)}r_j\right) + \gamma^l Q_{\phi'}^\pi(s_{l}', E_{\omega_1}(D_{\omega_2}(\pi(s_{l}')))).
\end{equation}
Please note that in Equation~\ref{td-targets-int} the routine used to evaluate the target Q-network $Q_{\phi'}^\pi$ at state $s_{l}'$ is not simply the routine selected by the policy $\pi(s_{l}')$, but rather its `auto-encoded' version $E_{\omega_1}(D_{\omega_2}(\pi(s_{l}')))$. The necessity of this comes from the fact that the Q-function is always updated with routines obtained by encoding action sequences sampled from the replay buffer distribution $E_{\omega_1}(a_{1:l})$, which, in practice, can differ from the distribution of routines outputted by the actor. 
Thus, the additional `auto-encoding' ensures that the inputs of the target Q-function also come from encoded action sequences, empirically providing further stabilization. 

This update can be performed efficiently with respect to all routines $n^{a_{1:L}}$ in parallel, making use of matrix multiplication. We provide pseudocode in Section~\ref{app:pseudo} of the Appendix.

\textbf{Learning a policy with routines.} Given an informative Q-function over our routine space, we can optimize a parameterized policy $\pi_\theta$ to maximize its expected output. Thus, we specify an objective similar to Equation~\ref{pi_pg_obj}, optimizing $\theta$ over the Q value estimates from on-policy routines. 
Namely, we maximize the routine policy gradient objective $J_\pi$:
\begin{equation}
\label{pi_pgi_obj}
\argmax_\theta\E_{s\sim B, n\sim \pi_\theta(\cdot|s)}\left[Q_\phi^{\pi_\theta} (s, E_{\omega_1}(D_{\omega_2}(n)))\right].
\end{equation}
As in Equation \ref{td-targets-int}, the output of the policy is `auto-encoded' before being evaluated by $Q_\phi^{\pi_\theta}$, bringing the distribution of evaluated routines closer to the one used to learn $Q_\phi^{\pi_\theta}$. The processing performed to compute all four optimization objectives is summarized in Figure \ref{figure:optimization-flow}.

\textbf{Building a maximally useful routine space.} In order to build a routine space that facilitates solving the target task, we utilize the training signal from the Q-function and policy optimizations to learn $E_{\omega_2}$ and $D_{\omega_1}$. 
Particularly, the routine encoder, $E_{\omega_2}$, is trained together with the Q-function to minimize the temporal difference objective $J_Q$ from Equation~\ref{qi_obj}. During this step, we also decode the routine representations computed from the action sequences in the replay buffer, $E_{\omega_1}(a_{1:l})$, to optimize the length consistency objective $J_{lc}$ from Equation~\ref{lc_obj}.
\begin{equation}
\label{enc_q_obj}
\argmin_{\omega} J_Q + J_{lc}.
\end{equation}
This optimization has the effect of building a routine space that facilitates the Q-function $Q_\phi^{\pi_\theta}$ to accurately represent the true Q values of different sets of action sequences. In particular, the routine encoder $E_{\omega_1}$ will be pushed to map action sequences with different Q values (thus, different effects towards achieving the task's objective) to different routine representations, facilitating $Q_\phi^{\pi_\theta}$ to minimize the TD-loss. Moreover, $E_{\omega_1}$ will be pushed to map the routine representation of unseen action sequences closer to the routine representation of other seen action sequences accomplishing the same task objective, for which $Q_\phi^{\pi_\theta}$ already minimizes the TD-loss. 

Similarly, the routine decoder is trained together with the policy to maximize the policy gradient objective 
$J_\pi$ from Equation~\ref{pi_pgi_obj}. 
We also utilize the samples from $n\sim \pi_\theta(\cdot|s)$ to jointly optimize the many-to-one consistency objective $J_{mto}$ with $E_{\omega_1}(D_{\omega_2}(i)))$ at minimal extra cost:
\begin{equation}
\label{dec_pi_obj}
\argmax_{\omega_2} J_\pi - J_{mto}.
\end{equation}
This optimization has the effect of building a routine space that facilitates the policy $\pi_\theta$ to output routines for which the decoded actions maximize our estimate of $Q_\phi^{\pi_\theta}$. Note that we do not optimize the routine encoder $E_{\omega_1}$ with respect to this objective, since doing so would encourage the `auto-encoded' routine $E_{\omega_1}(D_{\omega_2}(n))$ to diverge from the original routine $n$, in order to maximize Eq.~\ref{pi_pgi_obj}, without such change being necessarily reflected in the decoded action sequence from $D_{\omega_2}(n)$. Therefore, 
we avoid the mappings between the routine and action spaces changing only to adversarially \textit{fool} the Q-function $Q_\phi^{\pi_\theta}$ towards outputting high values.

\subsection{Advantages of Utilizing Routines}
\label{routine-advantage}
There are multiple advantages in learning an agent that reasons with routines rather than actions. Our framework increases the agent's expressivity, allowing it to output behavior executed at different timescales. As also empirically observed in prior works \citep{actreps_mot2, tempExt0}, the introduction of temporally-extended structured behavior appears to yield non-trivial exploration benefits in different tasks. Practically, this freedom also enables to lower the amount of reasoning the agent needs to perform to complete a particular episode since its policy can learn to select routines spanning multiple time-steps when a very fine level of control is not needed. This has non-trivial implications for speeding up experience collection, facilitating applications for time-sensitive real-world applications, and better scaling to expensive policy models such as neural network ensembles.

Another kind of advantage comes from the optimization of the reinforcement learning objectives in the routine framework. Reinforcement learning algorithms appear to be sensible to the granularity of their action spaces. For example, different action repeats have concrete effects on the agent's performance in different environments \cite{frameskip-importance}, even motivating some recent works to directly learn this parameter \citep{actpers0, actpers1}. Additionally, techniques such as learning Q-functions with multi-step targets often lead to faster learning by speeding up reward propagation \cite{sutton-reinforcement-n-step} and have been widely adopted in many state-of-the-art algorithms \cite{A3c, rainbow, d4pg}. By enabling agents to reason and learn at different frequencies concurrently, the routine framework provides a way to exploit faster reward propagation, without necessarily increasing the variance in the Q-function TD-targets. We perform experiments to better understand and decouple the exploration and learning benefits provided by our framework in Section~\ref{app:routineanalysis} of the Appendix.

Hence, there are several characteristics of the routine framework, differentiating it from prior works. Our framework does not rely on external heuristics such as state representations \citep{FuN, HIRO} or transitions information \citep{traj-embeddings-auto-encoder, actseq-rep-dynamics-aware-emb} to recover higher-level abstractions of behavior. Instead, our end-to-end approach should recover a routine space tailored to maximize the performance of each individual task, with the potential of mutating to facilitate different learning stages. Moreover, routines can represent arbitrarily complex distributions of action sequences, thus providing increased flexibility as compared to algorithms reasoning with action repetitions \citep{action-reps-discrete, action-reps-policy-factor, act-rep-cr}. Additionally, as described in this section, the computation of routines is specifically well suited for off-policy reinforcement learning, enabling to naturally harness the benefits of faster for reward propagation without its major downsides,

%% file: sections/5Algorithm.tex
\section{Practical Algorithms}

\label{sec:algorithms}

We integrate the routine framework and optimization procedures specified in Section~\ref{sec:method} with additional practices from two state-of-the-art off-policy reinforcement learning algorithms, described in Section \ref{advanced-pga}. 

\subsection{Integration with TD3} 

We first describe making use of the routine framework and additional models with 
\textit{TD3} \cite{td3}. 
Since \textit{TD3} aims to learn a deterministic policy, incorporating routines simply amounts to replacing the TD-loss and policy gradient loss with their routine counterparts, augmented by the many-to-one and length consistency objectives as specified in Section~\ref{end-to-end-opt}. 

Utilizing a routine decoder also allows injecting Gaussian exploration noise on two levels: the action level and the routine level. Injecting noise at the routine level provides \textit{TD3} means of using additional state-dependent exploration. In our experiments, we use a combination of these two types of noise which appears to yield the best results.  We investigate the effects of each distinct type of noise on performance in Section~\ref{app:routineanalysis} of the Appendix.

\subsection{Integration with SAC}
\label{td3-int}
We further describe integrating the routine framework with \textit{SAC} 
\cite{sac-alg}. Unlike \textit{TD3}, \textit{SAC} proposes to optimize an augmented maximum entropy objective with a stochastic policy outputting the parameters of an independent Gaussian distribution over actions. When reasoning with routines, we optimize this augmented objective by making use of a `Gaussian' routine decoder and a deterministic policy. 
Particularly, the routine decoder now outputs $L$ vectors for both the means $\mathbf{\mu}^n_{1:L}$ and standard deviations $\mathbf{\sigma}^n_{1:L}$ of independent multivariate Gaussian distributions, together with early termination probabilities $e^n_{1:L-1}$: $D_{\omega_1}: N \mapsto (\mathbb{R}^{2 \times a \times L}, \mathbb{R}^{L-1})$. To sample an action sequence from a routine $n$, we first sample the terminating index $l$, as in Equation~\ref{seg-len-prob}, to obtain the relevant vector means and standard deviations, $\mathbf{\mu}^n_{1:l}$ and $\mathbf{\sigma}^n_{1:l}$: 
\begin{align}
    p(\mathbf{\mu}^n_{1:l}, \mathbf{\sigma}^n_{1:l}|n) = e^n_l &\prod_{j=1}^{l-1} (1-e^n_j) \text{ for }  1\leq l<L, \nonumber \\
    p(\mathbf{\mu}^n_{1:L}, \mathbf{\sigma}^n_{1:L}|n) = &\prod_{j=1}^{L-1} (1-e^n_j).
    \label{seg-len-stoc-prob}
\end{align}
Subsequently, we use the relevant vector means and standard deviations to sample actions with probabilities $p(a_{1:l}|\mathbf{\mu}^n_{1:l}, \mathbf{\sigma}^n_{1:l})$ given by an independent multivariate Gaussian distribution:
\begin{equation}
    a^n_{1:l} \sim N(\mathbf{\mu}^n_{1:l}, (\mathbf{\sigma}^n_{1:l})^2).
\end{equation}

The TD-targets in \textit{SAC} are computed by augmenting the traditional backups with a discounted term evaluating the policy's entropy at the next state: $r + \gamma E_{a'\sim \pi(\cdot|s')}[Q^\pi_{\phi'}(s', a') - \pi(a'|s')]$. Similarly, we also augment the routine Q-function TD-targets from Equation~\ref{td-targets-int} with a discounted term, evaluating the routine decoder's entropy for the policy routine at the relevant subsequent state, $n'_l=\pi(s'_l)$. Particularly, the TD-targets for a routine recovered from an action sequence of length $l$ in the replay buffer are:
\begin{multline}
y^{SAC}_l = \sum^l_{j=1}\left(\gamma^{(j-1)}r_j\right) + \gamma^l \E_{a'_{1:l'}}
[Q_{\phi'}^\pi(s_{l}', E_{\omega_1}(a'_{1:l'})) + \\ 
-log(p(a_{1:l'}'|n'_l))].
\end{multline}
Policy optimization in \textit{SAC} is also similarly performed by augmenting the Q function output with a term incentivizing the policy's entropy: $\E_{s\sim B, a\sim \pi_\theta(\cdot|s)}\left[Q_\phi^{\pi_\theta} (s,a) - \pi_\theta(a|s) \right].$ Hence, we again convert this term to incentivize the routine decoder's entropy in decoding the policy routine at sampled states $n=\pi(s)$. This leads to a new policy optimization objective:
\begin{equation}
\label{pi_pgi_obj_sac}
J_\pi^{SAC}=\E_{B, a_{1:l}}\left[Q_\phi^{\pi_\theta} (s, E_{\omega_1}(a_{1:l})) - log(p(a_{1:l} |n)))\right].
\end{equation}
Unlike our \textit{TD3} integration, we do not explore adding stochasticity at the routine level from the policy's output. Further details regarding our integration with \textit{TD3} and \textit{SAC} are provided in Section~\ref{app:integration} of the Appendix.

%% file: sections/6Experiments.tex
\section{Experimental Results}
\label{sec-exp}

In this section, we provide an evaluation of the proposed routine framework utilizing the DeepMind Control Suite \cite{dmc}. We compare the performance of different algorithms as a function of the total number of observations experienced, with each epoch corresponding to experiencing $10,000$ observations.  
We repeat each experiment ten times for the main \textit{Performance Analysis} and five times for the other ablations, providing both the mean and standard deviation across the runs.
 We record five evaluation runs at the end of each epoch and collect both the cumulative returns and the total number of times that routine-based agents need to query their policy to complete an episode. Having to query the policy a lower number of times provides concrete improvements to the agent's computational efficiency, having implications in speeding up the process of experience collection and facilitating actuation. 

\begin{figure}
  \centering
  \includegraphics[width=0.95\columnwidth]{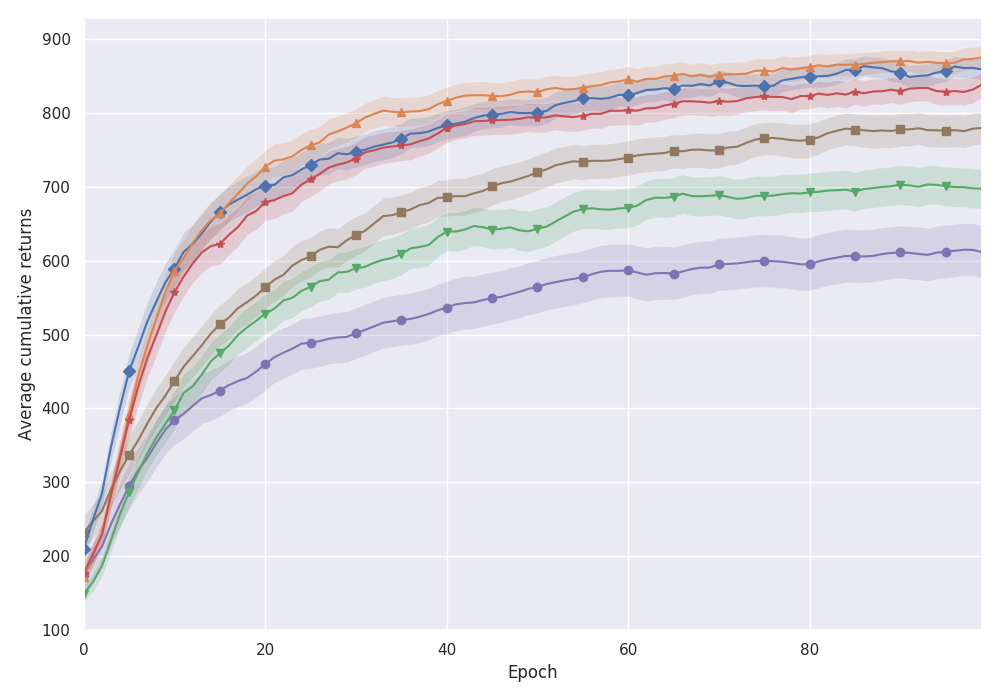}
  \includegraphics[width=0.95\columnwidth]{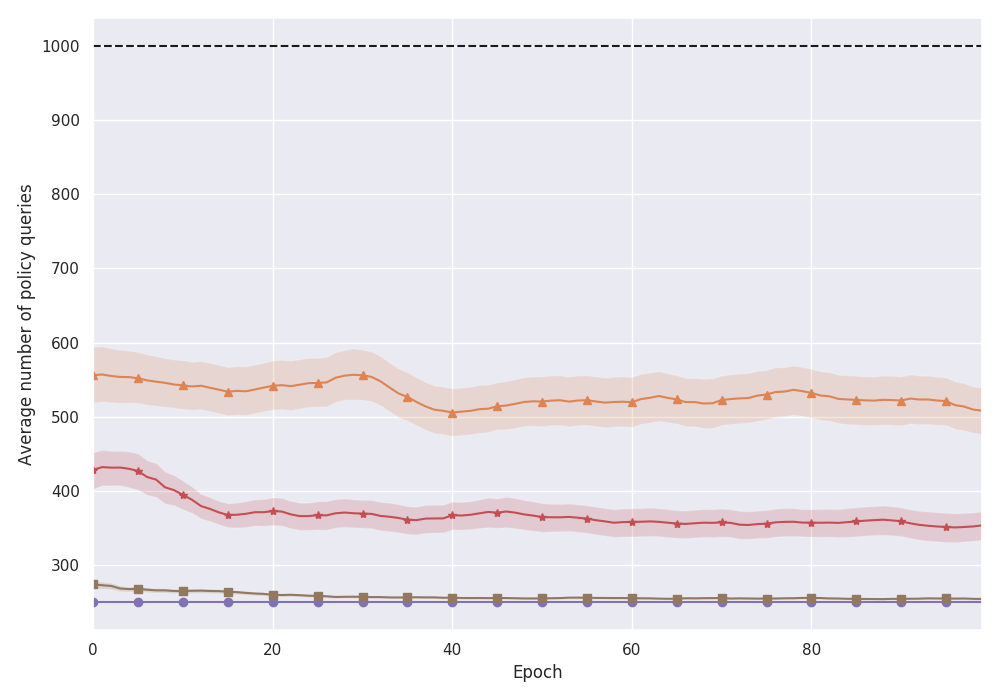}
  \includegraphics[width=0.9\columnwidth]{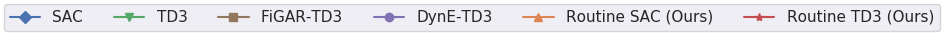}

  \caption{Average cumulative returns (Top) and number of policy queries (Bottom) across all fourteen evaluated environments, as a function of the epoch number. We repeat each experiment ten times and show both mean and standard deviations across all runs.}
  \label{figure:ave-perf}

\end{figure}

\begin{figure}
  \centering
  \includegraphics[width=0.95\columnwidth]{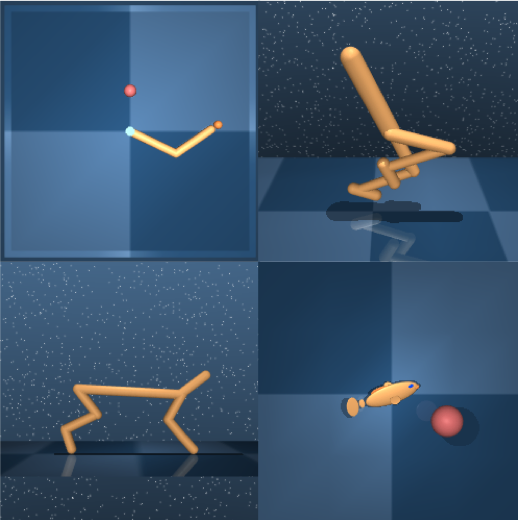}

  \caption{Visualization of the environments used in the expressivity analysis, namely, \textit{Reacher}, \textit{Walker}, \textit{Cheetah} and \textit{Fish}.}
  \label{figure:vis-envs}

\end{figure}

\begin{table}[t]
\caption{Average cumulative returns and number of policy queries (in brackets) over the last 10 epochs of training.} \label{tab:routine-L}
\vskip 0.15in
\tabcolsep=0.04cm
\begin{center}

\adjustbox{max width=0.98\columnwidth}{

\begin{tabular}{@{}lcccc@{}}
\toprule
                              & \textit{Reacher hard}       & \textit{Walker walk}        & \textit{Cheetah run}         & \textit{Fish swim}            \\ \cmidrule(l){2-5} 
\textit{TD3}                  & $962\pm 10$ ($1000$)        & $870\pm 95$ ($1000$)        & $862\pm 17$ ($1000$)         & $142\pm 81$ ($1000$)          \\
\textit{Routine TD3} $(L=2)$  & $778\pm 379$ ($994$)        & $927\pm 82$ ($855$)         & $817\pm 134$ ($619$)         & $444\pm 52$ ($621$)           \\
\textit{Routine TD3} $(L=4)$  & $\mathbf{975\pm 4}$ ($471$) & $\mathbf{975\pm 2}$ ($704$) & $\mathbf{880\pm 23}$ ($338$) & $745\pm 38$ ($315$)           \\
\textit{Routine TD3} $(L=8)$  & $967\pm 14$ ($336$)         & $969\pm 4$ ($194$)          & $856\pm 30$ ($243$)          & $797\pm 30$ ($189$)           \\
\textit{Routine TD3} $(L=16)$ & $963\pm 14$ ($357$)         & $924\pm 80$ ($203$)         & $725\pm 78$ ($195$)          & $\mathbf{807\pm 9}$ ($177$)   \\ \midrule
\textit{SAC}                  & $956\pm 13$ ($1000$)        & $968\pm 4$ ($1000$)         & $844\pm 31$ ($1000$)         & $530\pm 33$ ($1000$)          \\
\textit{Routine SAC} $(L=2)$  & $972\pm 6$ ($809$)          & $927\pm 82$ ($994$)         & $829\pm 27$ ($756$)          & $281\pm 65$ ($779$)           \\
\textit{Routine SAC} $(L=4)$  & $969\pm 10$ ($315$)         & $\mathbf{969\pm 1}$ ($495$) & $846\pm 15$ ($529$)          & $580\pm 140$ ($765$)          \\
\textit{Routine SAC} $(L=8)$  & $\mathbf{973\pm 3}$ ($322$) & $\mathbf{969\pm 4}$ ($296$) & $\mathbf{866\pm 14}$ ($443$) & $\mathbf{614\pm 185}$ ($307$) \\
\textit{Routine SAC} $(L=16)$ & $954\pm 30$ ($235$)         & $892\pm 73$ ($231$)         & $793\pm 49$ ($390$)          & $527\pm 282$ ($167$)          \\ \bottomrule
\end{tabular}
}
\end{center}
\end{table}

\subsection{Performance Analysis}
\label{perf-ana}
Firstly, to understand the effectiveness of the routine framework, we evaluate the novel routine off-policy algorithms specified in Section~\ref{sec:algorithms} in fourteen distinct tasks. For these experiments, we fix the maximum routine length to $L=4$. We provide all other hyper-parameters used by our algorithms in Section~\ref{app:algparams} of the Appendix. We compare their performance against the state-of-the-art baseline off-policy algorithms from which they are derived, and two algorithms from recent works that extend \textit{TD3} through temporal abstractions:
\begin{itemize}
    \item \textit{FiGAR TD3} - Modified version of the \textit{FiGAR DDPG} algorithm described by \citet{action-reps-policy-factor}, tuned to obtain improved performance in the DeepMind Control Suite environments. This algorithm parameterizes a policy outputting an action and a probability distribution over action repetitions. We provide details of our implementation in Section~\ref{app:figar-desc} of the Appendix.
    \item \textit{DynE TD3} - Algorithm proposed by \citet{actseq-rep-dynamics-aware-emb} that builds action-sequence embeddings using state information. We adapted the authors' original implementation.
\end{itemize}

In Figure~\ref{figure:ave-perf} we provide visualizations for both the average returns and the average number of policy evaluations per episode across all fourteen tasks as a function of the number of epochs. Note that for each task in the DeepMind Control Suite the obtainable cumulative returns are always in the range $[0, 1000]$. %
From the obtained results, we show that integrating the routine framework improves both examined state-of-the-art algorithms. Particularly, from the integration of routines with \textit{TD3} we observe substantial gains both in terms of final performance and convergence speed. %
The gains in final performance appear to be more significant in the harder exploration environments, indicating that the routine frameworks and the two levels noise injection described in Section \ref{td3-int} concretely help exploration. While \textit{SAC} already achieves high final performance on the DeepMind Control Suite, incorporating routines still provides improvements in terms of convergence speed and stability, showing the effectiveness of our framework. 

Our \textit{Routine TD3} algorithm also convincingly outperforms both \textit{FiGAR TD3} and \textit{DynE TD3}, especially in some of the more complex locomotion tasks in the \textit{Cheetah} and \textit{Walker} environments. We believe one of the reasons for this gap is that \textit{FiGAR} and \textit{DynE} 
introduce temporal abstractions with representations that are either fixed or learned from some particular heuristic. Hence, while these representations might work well for particular tasks, they struggle to be generally effective for a large suite of diverse tasks. In contrast, routines build a representation space optimized end-to-end to facilitate off-policy learning for any specific underlying task, which we believe is one of the main reasons for their effectiveness on diverse problems. Additionally, both \textit{FiGAR} and \textit{DynE} cannot model high-level behavior over a full set of arbitrary action sequences, lacking the expressiveness of our framework.

Both algorithms that make use of the routine framework required 
 around half the number of policy evaluations per episode, considerably improving efficiency from their baselines. Interestingly, we observe the routine version of \textit{TD3} to be marginally more efficient. Perhaps, this could be motivated by the lack of routine level noise applied to the routine version of \textit{SAC}, causing it to under-explore its routine space. Both \textit{FiGAR TD3} and \textit{DynE TD3} recover policies acting as infrequently as possible. However, this efficiency is partially symptomatic of their inability to recognize when to act with greater precision, being one of the reason for their lower performance in the harder environments.  We provide the full set of results obtained per environment, including the relative performance curves in Section~\ref{app:fullres} of the Appendix.

Having no incentives to output `long' routines, we would expect our optimizations to eventually converge to a policy that would exploit the finest level of control available to maximize performance. One way to counteract this would be to add small reasoning penalties to the reward each time the agent's policy is queried, making the agent aware of computational costs. While we do not make use of any such penalties, interestingly, our results show that the recovered routine policies do not converge to selecting exclusively `short' routines. Instead, they appear to converge to selecting mostly longer routines, with some shorter routines being performed in the harder environments in scenarios where the agent is physically unstable. We hypothesize this occurs because in the majority of states there are very small performance gains in reasoning at every step. At the same time, faster reward propagation and target policy smoothing can already provide some implicit stimulus to the Q values of `longer' routines in the stochastic optimization context of TD-learning.

\subsection{Expressivity Analysis}

\label{expr-ana}
Increasing the maximum routine length parameter $L$ should provide the agents with a greater potential for deployment efficiency and should make reward propagation occur at an even faster rate. On the other hand, we hypothesize that it would make the Q-function harder to optimize since it would need to learn accurate Q values for a routine space that represents a larger number of action sequences in $S^A$. Hence, we empirically examine the effects of increasing the maximum routine length parameter $L$ on both performance and deployment efficiency. We evaluate both novel routine algorithms with a maximum routine length parameter $L$ of 2, 4, 8, and 16 on the four environments shown in Figure \ref{figure:vis-envs}. Thus, we compare their performance and efficiency in contrast with the relative state-of-the-art baselines.

We summarize the results in Table \ref{tab:routine-L}, where we report the final cumulative returns and number of policy queries in the last ten epochs of training. We show the full performance curves in Section~\ref{app:fullres} of the Appendix. Particularly, the highest final cumulative returns in all but one environment are obtained for the maximum routine lengths of 4 and 8. Moreover, the efficiency of the recovered agents appears to visibly improve by increasing $L$, requiring less than a fifth of policy queries in some experiments with $L=8$ and $L=16$. However, by increasing the maximum routine length to 16 we start observing a tradeoff where we obtain further efficiency benefits but with some performance costs. Interestingly, we also observe that for some environments, setting $L=2$ leads to slightly worse performance both with respect to the other routine algorithms but also with respect to the non-routine baselines. We believe this is because, in such environments, the expressivity added by a routine framework that can represent only action sequences of length 2 is too limited to outweigh the downsides that the increased model complexity introduces.

%% file: sections/7Conclusion.tex
\section{Discussion and Future Work}

We proposed the routine framework, a novel approach to reinforcement learning that strives to learn an effective higher-dimensional `action space' named the routine space, achieving improved performance and deployment efficiency.  Each routine is trained end-to-end to represent a set of length-agnostic action sequences with equivalent effects for tackling a target task. Our framework is fully compatible with off-policy reinforcement learning and we demonstrate its effectiveness by successfully applying it to two state-of-the-art algorithms for continuous control. Empirically, we show that the novel resulting algorithms consistently achieve better final cumulative returns and convergence speed than their relative baselines and two recent alternative frameworks for reasoning with temporal abstractions. Moreover, the recovered agents can avoid reasoning at every step, requiring only a fraction of the policy queries to complete an episode, yielding non-trivial benefits for computational efficiency and real-world deployment. One natural extension for future works would be to apply the routine framework for algorithms working in discrete environments, to test the ubiquity of its applicability and effectiveness. Additionally, it would be interesting to analyze the structure and transferability of the routine spaces recovered in different tasks as a way to assess similarity and perform few-shot learning.

%% file: appsections/APseudocode.tex
\section{Routine TD-Loss Pseudocode}

\label{app:pseudo}

As mentioned in Section \ref{end-to-end-opt}, we can learn the routine Q-function by performing efficient TD-updates for routines of all lengths, making use of matrix multiplication. Particularly, starting with a sampled action sequence from the replay buffer $a_{1:L}$, we calculate the TD-loss for all routines $n^{a_1}, n^{a_1, a_2}, ..., n^{a_{1:L}}$ as described by the following pseudocode based on the \textit{TensorFlow} syntax:

\lstset{style=mypython}
\begin{lstlisting}
# D, E: Routine decoder and encoder network
# Q, Q_tar: Q and target Q networks
# Pi: Policy
# gamma: discount factor
# L: maximum routine length

# dim: L x L
#[[1,     1,        1, ...,           1,            1],
# [0, gamma,    gamma, ...,       gamma,        gamma],
# [0,     0, gamma**2, ...,    gamma**2,     gamma**2],
# [                    ...                           ],
# [0,     0,        0, ...,gamma**(L-2), gamma**(L-2)],
# [0,     0,        0, ...,           0, gamma**(L-1)]]
r_discounts = (band_part(ones((L, L)), 0, -1)
    *constant([[gamma**i] for i in range(L)])
    
# dim: L
#[gamma, gamma**2, ..., gamma**L]
next_q_discounts = constant(
    [gamma**i for i in range(1, L+1)])

# input dim: N x |s|, N x |a|, N x L x |s|,
#            N x L,   N x L
def routine_TD_loss(s, a, next_s, r, t):
    # on-policy routine from next states 
    # dim: N x L x |a| 
    next_n = Pi(next_s)
    
    # autoencoded on-policy routine
    # dim: N x L x |a|
    ae_next_n = D.get_routine(
        E.sample_actions(next_n))
    
    # targets
    # dim: N x L
    y = (matmul(r, r_discounts) + 
         t*next_q_discounts*Q_tar(next_s, ae_next_n))
    
    # routines from actions subsequences in a
    # dim N x |n|
    n = D.get_routine_subsequences(a)
    
    # tiled Q predictions
    # dim N x L
    q = Q(tile(reshape(s, [N, |s|, 1]), [1, 1, L]), n)
    
    # return average TD-loss and encoded routines
    return mean(square(q - y)), n
\end{lstlisting}

For further details, please refer to our shared implementation. 

%% file: appsections/BIntegration.tex
\section{Integration Details}

\label{app:integration}

\begin{algorithm}
    \caption{\textit{Routine off-policy framework}}
    \label{routine_alg}
    \begin{algorithmic}[1]
    \STATE Initialize $\pi_\theta, Q_{\phi_1}^{\pi_\theta}, Q_{\phi_2}^{\pi_\theta}, D_{\omega_1}, E_{\omega_2}.$
    \STATE Initialize $B_\pi \gets \emptyset$, $count\gets 0$
    \FOR{$i = 1, 2, ...,$}
        \STATE Observe $s$ from the environment
        \STATE Query policy to obtain $n\sim \pi_{\theta}(s)$
        \STATE Sample $a_1, a_2, ..., a_l \sim D_{\omega_1}(n)$
        \FOR{$a = a_1, a_2, ..., a_l$}
            \STATE Execute $a$ in the environment, collect $(s', r, t)$
            \STATE Store $B = B \cup (s, a, s', r, t)$
            \STATE Sample $b=\{(s, a_{1:L}, s_{1:L}', r_{1:L}, t_{1:L})_{|b|}\}\in B$
            \FOR{$m = 1, 2$}
                \STATE Approximate $J_Q$ and $J_{lc}$ with $b$
                \STATE Update $Q^{\pi_\theta}_{\phi_m}$ with $\nabla_{\phi_m}J_Q$
                \STATE Update $D_{\omega_1}$, $E_{\omega_2}$ with $\nabla_{\omega}(J_Q+J_{lc})$
            \ENDFOR
            \STATE Update $count\gets count + 1$
            \IF{$count \mod delay$ = 0}
                \STATE Approximate $J_\pi$ and $J_{mto}$ with $b$
                \STATE Update $\pi_\theta$ with $\nabla_{\theta}J_\pi$
                \STATE Update $D_{\omega_1}$ with $\nabla_{\omega_1}(J_\pi-J_{mto})$
            \ENDIF
        \ENDFOR
    \ENDFOR
    \end{algorithmic}
\end{algorithm}


In Algorithm \ref{routine_alg} we show the common optimization structure of our routine framework. Below, we further provide more details regarding the integration of our framework with \textit{TD3} \citep{td3} and \textit{SAC} \citep{sac}, including few conceptual dissimilarities with the original algorithms.

The \textit{Routine TD3} algorithm explores the environment by making use of independent Gaussian noise injected both at the routine and action levels. Moreover, when calculating the TD-loss for learning the routine Q-function, we chose to utilize the actual policy to obtain the next state target routine and avoid parameterizing a target policy as in the original algorithm. This choice did not appear to influence particularly the performance and was done with the purpose of simplification.

The \textit{Routine SAC} algorithm bases its implementation on the automatic temperature adjustment version of \textit{SAC} \cite{sac-alg}. Particularly, we keep the same original heuristic for the environment-specific action-selection entropy target of $-|a|$. However, when updating the temperature parameter $\alpha$, we still utilize this target against the decoder's `per-action entropy', rather than the overall entropy of chosen routines. Practically, the decoder's `per-action entropy' is simply calculated dividing the overall entropy from the decoded action sequence distribution by its recovered length. Additionally, we make use of delayed policy training, as in \textit{TD3}, updating our policy and target networks less frequently than the routine Q-function. 

%% file: appsections/CHyper.tex
\section{Algorithms Parameters}

\label{app:algparams}

In this section, we describe the hyper-parameters choices made for all the evaluated algorithms in Section \ref{sec-exp}.

For the \textit{TD3} and \textit{SAC} algorithms we utilized the parameters provided in the original implementations. We share most of the \textit{TD3} and \textit{SAC} parameters with our routine-based versions of these algorithms with only minor differences. For example, in \textit{TD3} we utilize a smaller target routine smoothing value of $0.1$ to regularize the auto-encoded version of the predicted next state routine. Additionally, in both routine versions of \textit{TD3} and \textit{SAC} we use 2-layer fully-connected networks with 256 hidden units for both policy and Q function models.

We utilized simple rules to select the dimensionality of the routine representations and keep the structure of the additional routine decoder and encoder models light and efficient, comprising only a few hundred additional parameters. Particularly, using the notation from Figure \ref{figure:enc-dec-models}, 
we let the routine space representation dimensionality be based on the original environment's action space dimensionality: $|n|=L\times |a|$. Additionally, we respectively set the first layer embeddings dimensionality to $|h|=2^{\left\lceil{log_2(|a|)}\right\rceil}$ and the aggregated representation dimensionalities to $|g|=L\times |h|$. As we wanted to evaluate the general applicability of our framework, we did not substantially tune the hyper-parameters of these models. Thus, we did not explore using any information bottleneck between the action sequences space $S^A$ and the routine space, but hypothesize this could yield even further efficiency improvements.

We list all the hyper-parameter choices in Table \ref{tab:hyper}.

\begin{table}[t]
\caption{Hyper-parameters used for the experimental evaluation.} \label{tab:hyper}
\vskip 0.15in
\tabcolsep=0.08cm
\begin{center}
\adjustbox{max width=0.98\columnwidth}{
\begin{tabular}{@{}lc@{}}
\toprule
\multicolumn{2}{c}{Shared parameters}                                                   \\ \midrule
buffer size $|B|$                           & $100000$                                  \\
batch size $|b|$                            & $256$                                     \\
minimum data to train                       & $1000$                                    \\
optimizer                                   & Adam                                      \\
learning rate                               & $0.001$                                   \\
optimizer $\beta_1$                                   & $0.9$                                     \\
policy delay                                & $2$                                       \\
discount $\gamma$                           & $0.99$                                    \\
polyak coefficient $\rho$                   & $0.995$                                   \\
policy/Q network hidden layers              & $2$                                       \\
policy/Q network hidden dimensionality      & $256$                                     \\
routine space dimensionality $|n|$          & $L\times |a|$                             \\
decoder/encoder hidden dimensionality $|h|$ & $2^{\left\lceil{log_2(|a|)}\right\rceil}$ \\
encoder aggregated dimensionality $|g|$     & $L\times |h|$                             \\
$J_{mto}$ coefficient                       & $1$                                       \\
$J_{lc}$ coefficient                        & $1$                                       \\ \midrule
\multicolumn{2}{c}{Routine \textit{TD3} parameters}                                     \\ \midrule
routine exploration noise                   & $0.2$                                     \\
action exploration noise                    & $0.1$                                     \\
target routine smoothing noise              & $0.1$                                     \\ \midrule
\multicolumn{2}{c}{Routine \textit{SAC} parameters}                                     \\ \midrule
starting entropy temperature $\alpha$       & $0.1$                                     \\
entropy temperature learning rate           & $0.0001$                                  \\
entropy temperature optimizer $\beta_1$               & $0.5$                                     \\ \bottomrule
\end{tabular}}
\end{center}
\end{table}

%% file: appsections/DFigarTD3.tex
\section{Implementation of Prior Algorithms}

\label{app:figar-desc}

To compare the routine framework with prior methods reasoning with action repetitions, we implemented the off-policy version of the \textit{FiGAR} algorithm by \citet{action-reps-policy-factor}, named \textit{FiGAR DDPG}. This algorithm works by parameterizing a policy outputting both an action and a probability distribution over a set of possible action repetitions. However, strictly following the implementations details and hyper-parameters described in the original paper yielded an algorithm which failed to learn meaningful behavior for the DeepMind Control Suite tasks. Thus, we implemented \textit{FiGAR TD3}, a new algorithm that extends \textit{FiGAR DDPG} by incorporating advances from \textit{TD3}, together with several additional practices to stabilize its optimization procedures.

Particularly, \textit{FiGAR TD3} makes use of double Q-learning, target policy smoothing, and delayed policy updates, as outlined in the paper by \citet{td3}. Additionally, we found two additional changes that played an even more significant role on performance. These consist in greatly reducing the range of possible action repetitions and augmenting the original experience collection procedure. Specifically, \textit{FiGAR DDPG} only records transitions in the replay buffer corresponding to the executed actions and repetitions. Instead, we augment the experience collection procedure by recording transitions after each environment step, relabeling the intermediate steps with the appropriate action repetitions. The original paper by \citet{action-reps-policy-factor} is also ambiguous on how the repetition values are logged in the replay buffer. We tried utilizing both the actor's normalized outputted logits and a one-hot representation, with the latter approach yielding substantially better results. We further modified most of the original hyper-parameters for performance, as shown Table~\ref{tab:figarparam}, for further details please refer to our shared implementation.

\begin{table}[t]
\caption{Hyper-parameters used for \textit{FiGAR TD3}} \label{tab:figarparam}
\vskip 0.15in
\tabcolsep=0.08cm
\begin{center}
\adjustbox{max width=0.98\columnwidth}{
\begin{tabular}{@{}lc@{}}
\toprule
\multicolumn{2}{c}{\textit{FiGAR-TD3} parameters} \\ \midrule
buffer size $|B|$                      & $100000$ \\
batch size $|b|$                       & $256$    \\
minimum data to train                  & $1000$   \\
optimizer                              & Adam     \\
learning rate                          & $0.001$  \\
optimizer $\beta_1$                    & $0.9$    \\
policy delay                           & $2$      \\
discount $\gamma$                      & $0.99$   \\
polyak coefficient $\rho$              & $0.995$  \\
policy/Q network hidden layers         & $2$      \\
policy/Q network hidden dimensionality & $256$    \\
action exploration noise               & $0.1$    \\
repetition exploration $\epsilon$      & $0.2$    \\
exploration $\epsilon$ annealing steps & $50000$  \\
target action smoothing noise          & $0.1$    \\ \bottomrule
\end{tabular}
}
\end{center}
\end{table}

%% file: appsections/DFullRes.tex
\section{Full Results}

\label{app:fullres}

In this section, we provide the per-environment experimental results for the proposed routine framework.

\begin{figure*}
  \centering
  \includegraphics[width=\textwidth]{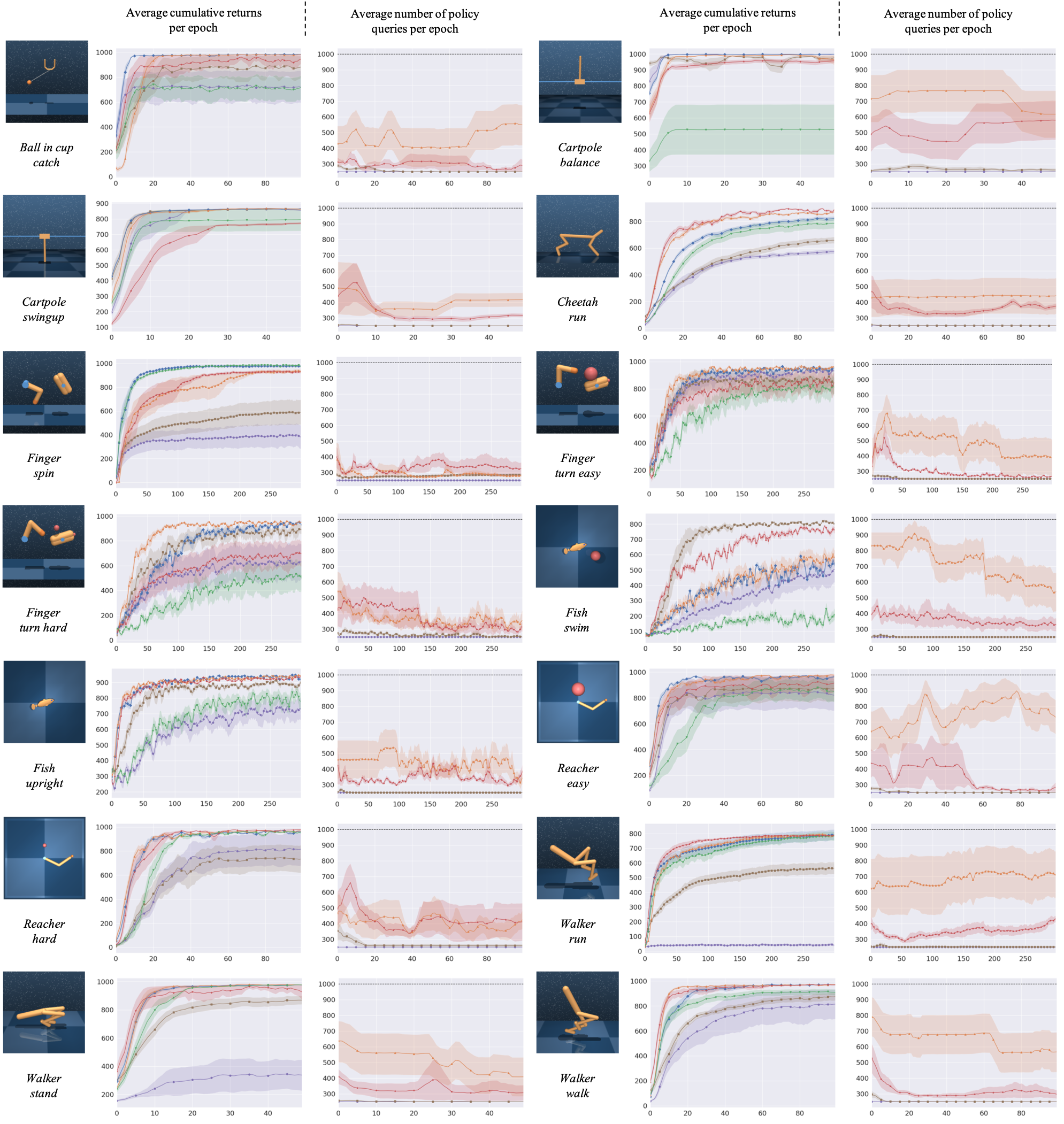}
  \includegraphics[width=\columnwidth]{figures/cr/cb/cb_perf_ana_legend.png}
  \caption{Average cumulative returns and number of policy queries in each of the fourteen different tested environments for the \textit{Performance Analysis} from Section \ref{perf-ana}. Particularly, we report the mean and the standard deviation of these quantities across ten different runs as a function of the epoch number.}
  \label{figure:all-exp-1}
\end{figure*}

In Figure \ref{figure:all-exp-1} we show the performance curves representing the average cumulative returns obtained and the average number of policy queries as a function of the epoch in each of the fourteen tested environments for the \textit{Performance Analysis} (from Section \ref{perf-ana}). We see the greatest performance gains of the routine framework occur for the \textit{TD3} algorithms in the harder exploration tasks. Overall, for the great majority of tasks, both routine versions of the examined algorithms provide improvements over their baselines. The average number of policy queries required to complete an episode appears to vary across the different environments. This can be seen as additional evidence that within the routine framework, agents do adaptively select routines of different lengths based on the granularity required to effectively solve a task. \textit{Routine TD3} also outperforms both \textit{FiGAR TD3} and \textit{DynE TD3}, which appear to particularly struggle in some of the more complex locomotion tasks in the \textit{Cheetah} and \textit{Walker} environments.

\begin{figure*}
  \centering
  \includegraphics[width=\textwidth]{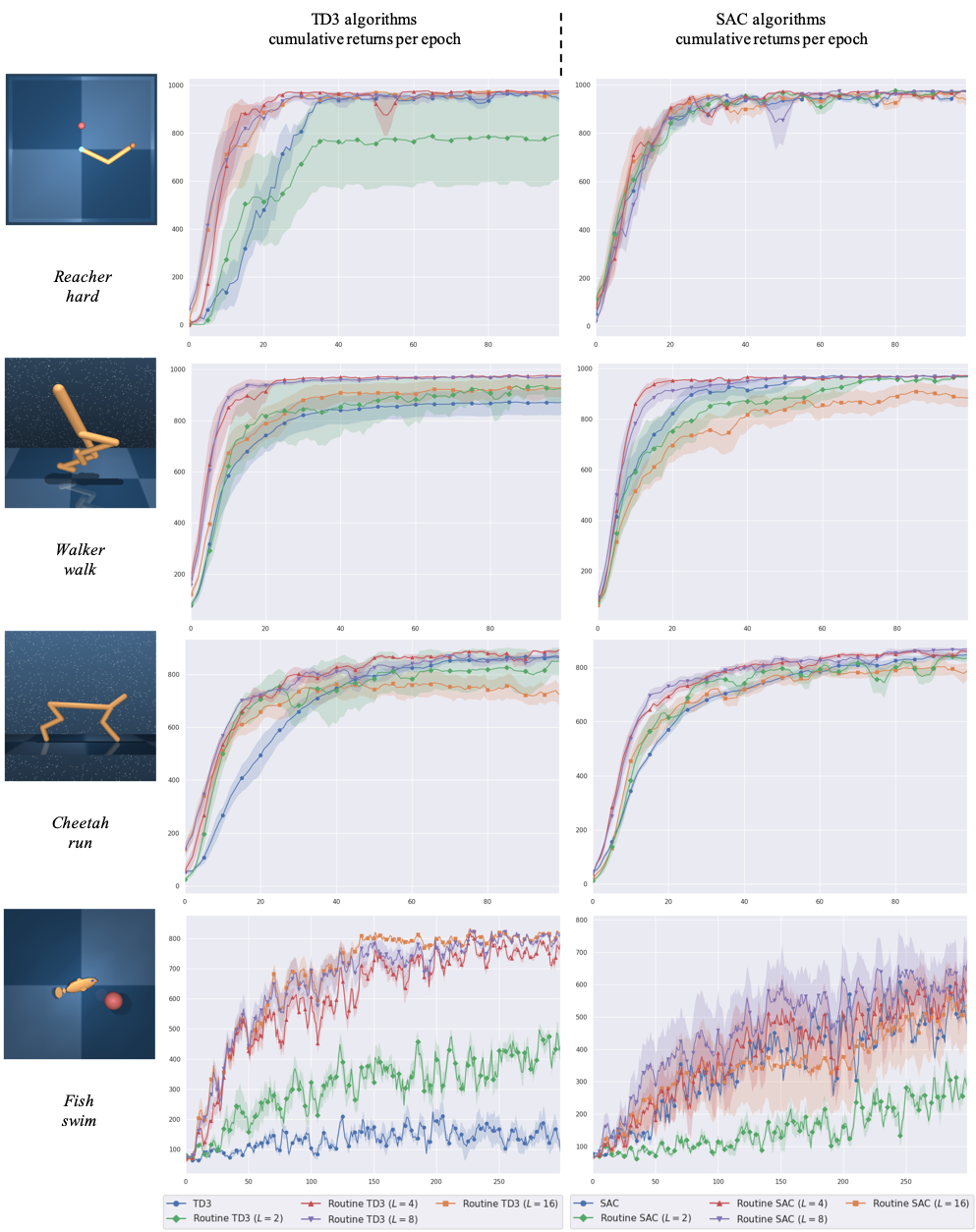}
  \vspace{-0.25cm}
  \caption{Cumulative returns in each of the four environments considered for the \textit{Expressivity Analysis} in Section \ref{expr-ana}. We show the performance curves for the original algorithms and the integration of the routine framework for \textit{TD3} (Left) and \textit{SAC} (Right).} 
  \label{figure:expr_res}
\end{figure*}

In Figure \ref{figure:expr_res} we provide the performance curves detailing the cumulative returns obtained by varying the maximum routine length $L$ to 2, 4, 8, and 16 for our \textit{Expressivity Analysis} (from Section \ref{expr-ana}). Overall, in terms of performance and stability, the best results are obtained by using a maximum routine length of 4 for our integration with \textit{TD3} and a maximum routine length of 8 for our integration with \textit{SAC}. These values appear to most optimally tradeoff the increased optimization complexity with the exploration, reward propagation and abstraction advantages provided by the routine framework.

%% file: appsections/FRoutineSpaceAnalysis.tex
\section{Routine Analysis}

\label{app:routineanalysis}

\begin{figure}
  \centering
  \includegraphics[width=0.95\columnwidth]{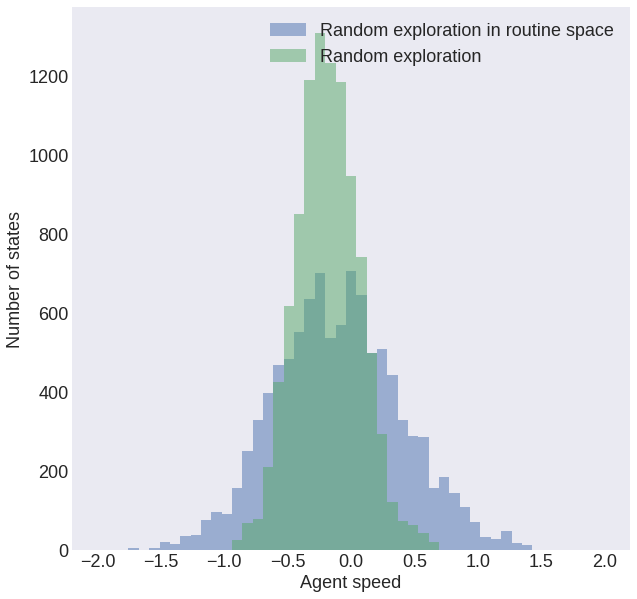}
  \caption{Number of states visited in ten episodes of experience collected in the \textit{Cheetah run} task, as classified by the internal Mujoco `speed' property (corresponding to the agent's velocity). We compare random uniform exploration by sampling from either the action space or the routine space.}
  \label{figure:abl-speed-expl}
\end{figure}

In this section, we provide a further analysis of the routine framework and its main components through additional ablations and visualizations. For the experiments in this section, we show the average performance of different algorithms and configurations obtained on the subset of four task introduced in Section~\ref{expr-ana}. 

\begin{figure}
  \centering
  \includegraphics[width=0.95\columnwidth]{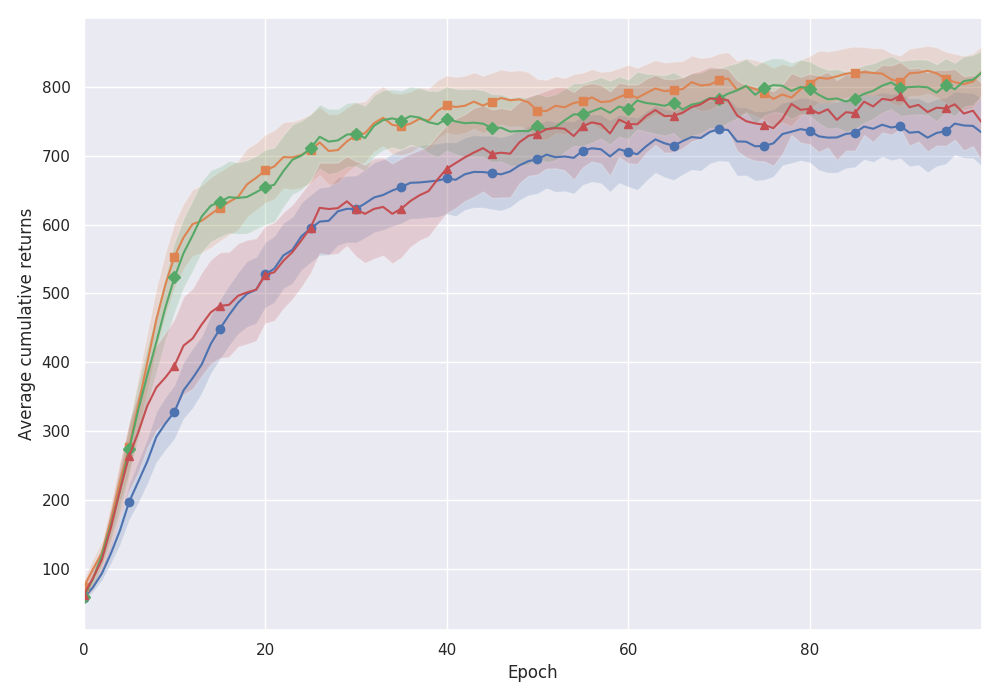}
  \includegraphics[width=0.9\columnwidth]{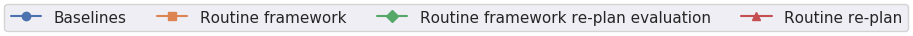}
  \caption{Average cumulative returns across the analyzed subset of tasks from decomposing the advantages of structured exploration from faster reward propagation. We average the performance of both \textit{SAC}- and \textit{TD3}-based algorithms for each of the different considered configurations of the routine framework. We repeat each experiment five times.}
  \label{figure:abl-replan}
\end{figure}

\subsection{Exploration and Learning Benefits of Routines}

As explained in Section~\ref{sec:method}, we hypothesize that the performance benefits observed from applying the routine framework to off-policy reinforcement learning algorithms come from both structured exploration and faster reward propagation. 

To reinforce the hypothesis that routines facilitate structured exploration, we compare the states encountered through action-based and routine-based exploration. Particularly, we consider the \textit{Cheetah run} task and collect different states from uniformly sampling either actions or routines. We categorize the states based on an internal Mujoco property named `speed', representing the velocity of the agent. We use this property as an indicative way of separating states corresponding to behavior with different effects on the underlying task. We show the results in Figure~\ref{figure:abl-speed-expl}, illustrating that routine-based exploration reaches states covering a significantly wider range of `speeds', validating our hypothesis.

We also perform additional ablation experiments aimed at decoupling the benefits of structured exploration from faster reward propagation. Particularly, we implement new versions of our routine algorithms which are forced to re-plan while acting, by selecting a new routine at every environment step and only executing its first action. Hence, these agents should still benefit from faster reward propagation during learning, but lose the hypothesized structured exploration benefits during experience collection.

We summarize the performance of the \textit{Routine re-plan} algorithms in Figure~\ref{figure:abl-replan}. For comparison, we use the performance obtained by the original routine algorithms both through standard execution and also matching the evaluation procedure of their re-planning counterparts, while still using full routines for experience collection. We average the performance of applying each considered setting of the routine framework to both \textit{SAC} and \textit{TD3} algorithms. The results show that the \textit{Routine re-plan} agents initially learn slower, yet, eventually clearly outperform the \textit{SAC} and \textit{TD3} baselines. Their performance also lags consistently behind the standard routine algorithms (under both evaluation schemes), reinforcing our hypothesis that our framework provides complementary benefits both regarding structured exploration and faster reward propagation.

\begin{figure}
  \centering
  \includegraphics[width=0.95\columnwidth]{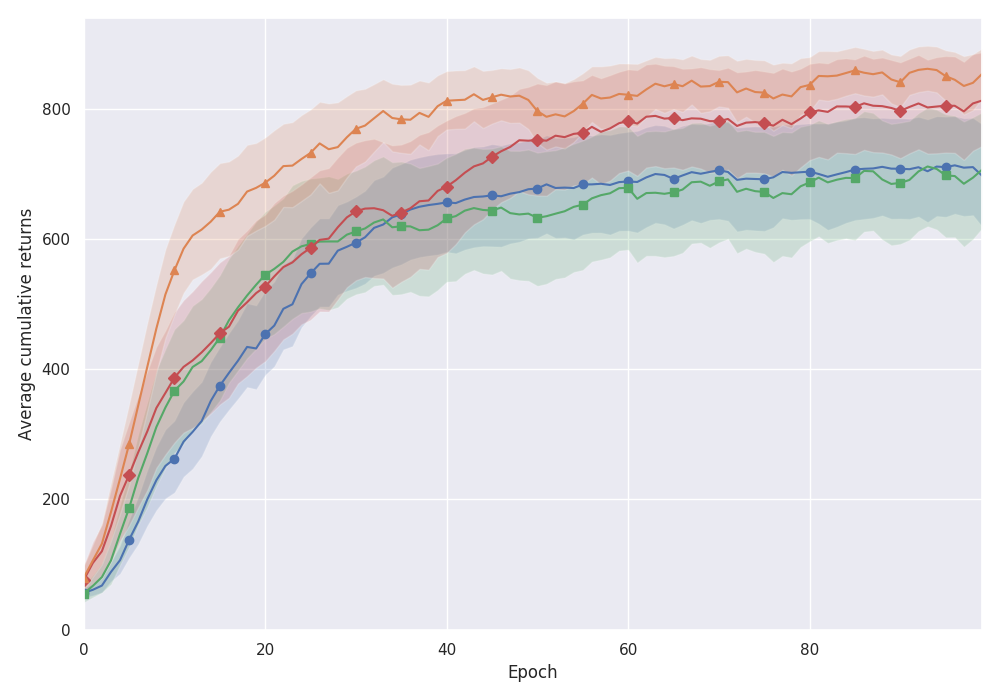}
  \includegraphics[width=0.9\columnwidth]{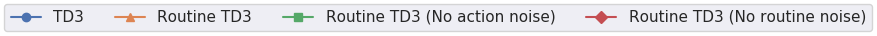}
  \caption{Average cumulative returns across the analyzed subset of tasks from removing either action space or routine space noise from \textit{Routine TD3}. We repeat each experiment five times.}
  \label{figure:abl-noise}
\end{figure}

\subsection{Effects of Routine Space Noise}

We analyze the effects of removing either the routine space or action space exploration noise from the \textit{Routine TD3} algorithm. We summarize the results in Figure~\ref{figure:abl-noise}. Both types of noise appear to have a positive impact on both final performance and learning speed. Action space noise appears to be a crucial component in exploration throughout learning and disabling it makes \textit{Routine TD3} converge to significantly worse policies. Routine space noise appears to have a greater effect on exploration early on, affecting more prominently the algorithm's learning speed.

\begin{figure*}
  \centering
  \includegraphics[width=0.95\textwidth]{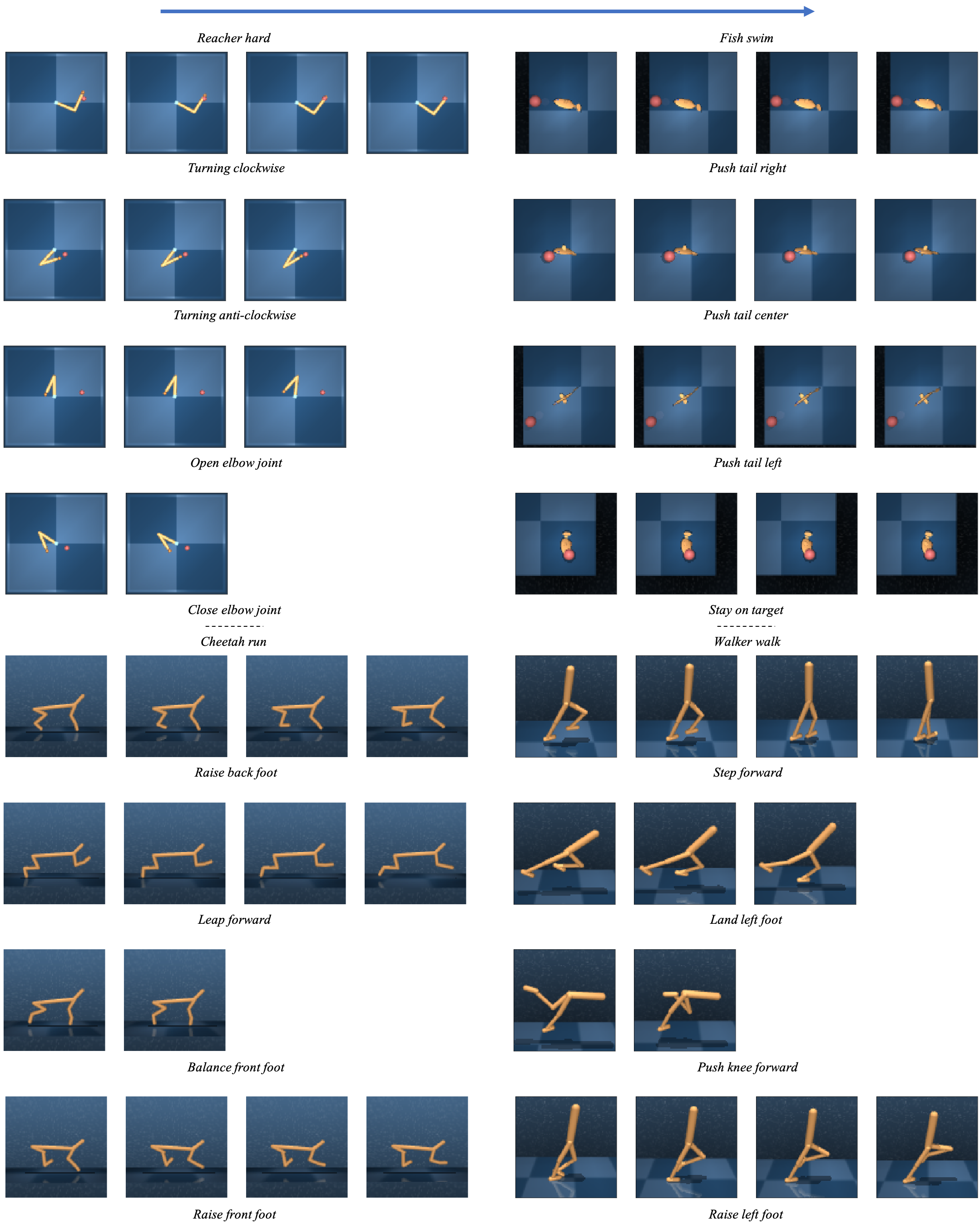}
  \caption{Visualizations of random routines by rendering the environments after executing each action sampled by the routine decoder. We assign each a semantic label based on their observed effects.}
  \label{figure:rout-vis}
\end{figure*}

\subsection{Routines Visualization}

To understand what kinds of behavior are encoded in routines, we collect visualizations by rendering the considered environments after performing each of the actions sampled by the routine decoder. We show these renderings in Figure \ref{figure:rout-vis}, assigning them corresponding semantic labels. Specifically, different routines appear to perform simple behaviors that can be reused effectively in multiple situations, allowing the policy to reason with higher-level abstractions. For example, in the \textit{Fish swim} task, different routines correspond to moving the agent's tail in different directions and to different extents, allowing the agent to maneuver towards any target.

%% file: main.bbl
\begin{thebibliography}{48}
\providecommand{\natexlab}[1]{#1}
\providecommand{\url}[1]{\texttt{#1}}
\expandafter\ifx\csname urlstyle\endcsname\relax
  \providecommand{\doi}[1]{doi: #1}\else
  \providecommand{\doi}{doi: \begingroup \urlstyle{rm}\Url}\fi

\bibitem[Bacon et~al.(2017)Bacon, Harb, and Precup]{option-critic}
Bacon, P.-L., Harb, J., and Precup, D.
\newblock The option-critic architecture.
\newblock In \emph{Proceedings of the AAAI Conference on Artificial
  Intelligence}, volume~31, 2017.

\bibitem[Bahl et~al.(2020)Bahl, Mukadam, Gupta, and Pathak]{NDP}
Bahl, S., Mukadam, M., Gupta, A., and Pathak, D.
\newblock Neural dynamic policies for end-to-end sensorimotor learning.
\newblock \emph{arXiv preprint arXiv:2012.02788}, 2020.

\bibitem[Barth-Maron et~al.(2018)Barth-Maron, Hoffman, Budden, Dabney, Horgan,
  Tb, Muldal, Heess, and Lillicrap]{d4pg}
Barth-Maron, G., Hoffman, M.~W., Budden, D., Dabney, W., Horgan, D., Tb, D.,
  Muldal, A., Heess, N., and Lillicrap, T.
\newblock Distributed distributional deterministic policy gradients.
\newblock \emph{arXiv preprint arXiv:1804.08617}, 2018.

\bibitem[Biedenkapp et~al.(2020)Biedenkapp, Rajan, Hutter, and
  Lindauer]{act-rep-cr}
Biedenkapp, A., Rajan, R., Hutter, F., and Lindauer, M.
\newblock Towards temporl: Learning when to act.
\newblock In \emph{Workshop on Inductive Biases, Invariances and Generalization
  in Reinforcement Learning (BIG@ ICML’20)}, 2020.

\bibitem[Braylan et~al.(2015)Braylan, Hollenbeck, Meyerson, and
  Miikkulainen]{frameskip-importance}
Braylan, A., Hollenbeck, M., Meyerson, E., and Miikkulainen, R.
\newblock Frame skip is a powerful parameter for learning to play atari.
\newblock In \emph{Workshops at the Twenty-Ninth AAAI Conference on Artificial
  Intelligence}, 2015.

\bibitem[Chandak et~al.(2019)Chandak, Theocharous, Kostas, Jordan, and
  Thomas]{single-action-rep-discrete-2}
Chandak, Y., Theocharous, G., Kostas, J., Jordan, S., and Thomas, P.~S.
\newblock Learning action representations for reinforcement learning.
\newblock \emph{arXiv preprint arXiv:1902.00183}, 2019.

\bibitem[Co-Reyes et~al.(2018)Co-Reyes, Liu, Gupta, Eysenbach, Abbeel, and
  Levine]{traj-embeddings-auto-encoder}
Co-Reyes, J.~D., Liu, Y., Gupta, A., Eysenbach, B., Abbeel, P., and Levine, S.
\newblock Self-consistent trajectory autoencoder: Hierarchical reinforcement
  learning with trajectory embeddings.
\newblock \emph{arXiv preprint arXiv:1806.02813}, 2018.

\bibitem[Dabney et~al.(2020)Dabney, Ostrovski, and Barreto]{tempExt0}
Dabney, W., Ostrovski, G., and Barreto, A.
\newblock Temporally-extended $\{\backslash$epsilon$\}$-greedy exploration.
\newblock \emph{arXiv preprint arXiv:2006.01782}, 2020.

\bibitem[Dayan \& Hinton(1993)Dayan and Hinton]{opt-sub-1}
Dayan, P. and Hinton, G.~E.
\newblock Feudal reinforcement learning.
\newblock In Hanson, S., Cowan, J., and Giles, C. (eds.), \emph{Advances in
  Neural Information Processing Systems}, volume~5, pp.\  271--278.
  Morgan-Kaufmann, 1993.
\newblock URL
  \url{https://proceedings.neurips.cc/paper/1992/file/d14220ee66aeec73c49038385428ec4c-Paper.pdf}.

\bibitem[Dietterich(2000)]{opt-sub-2}
Dietterich, T.~G.
\newblock Hierarchical reinforcement learning with the maxq value function
  decomposition.
\newblock \emph{Journal of artificial intelligence research}, 13:\penalty0
  227--303, 2000.

\bibitem[Dulac-Arnold et~al.(2015)Dulac-Arnold, Evans, van Hasselt, Sunehag,
  Lillicrap, Hunt, Mann, Weber, Degris, and
  Coppin]{single-action-rep-discrete-1}
Dulac-Arnold, G., Evans, R., van Hasselt, H., Sunehag, P., Lillicrap, T., Hunt,
  J., Mann, T., Weber, T., Degris, T., and Coppin, B.
\newblock Deep reinforcement learning in large discrete action spaces.
\newblock \emph{arXiv preprint arXiv:1512.07679}, 2015.

\bibitem[Dulac-Arnold et~al.(2019)Dulac-Arnold, Mankowitz, and
  Hester]{challenges-rl}
Dulac-Arnold, G., Mankowitz, D., and Hester, T.
\newblock Challenges of real-world reinforcement learning.
\newblock \emph{arXiv preprint arXiv:1904.12901}, 2019.

\bibitem[Florensa et~al.(2017)Florensa, Duan, and
  Abbeel]{hierarchical-subgoals-small-prior}
Florensa, C., Duan, Y., and Abbeel, P.
\newblock Stochastic neural networks for hierarchical reinforcement learning.
\newblock \emph{arXiv preprint arXiv:1704.03012}, 2017.

\bibitem[Fujimoto et~al.(2018)Fujimoto, Van~Hoof, and Meger]{td3}
Fujimoto, S., Van~Hoof, H., and Meger, D.
\newblock Addressing function approximation error in actor-critic methods.
\newblock \emph{arXiv preprint arXiv:1802.09477}, 2018.

\bibitem[Gregor et~al.(2016)Gregor, Rezende, and
  Wierstra]{hierarchical-subgoals-intrinsic-2}
Gregor, K., Rezende, D.~J., and Wierstra, D.
\newblock Variational intrinsic control.
\newblock \emph{arXiv preprint arXiv:1611.07507}, 2016.

\bibitem[Haarnoja et~al.(2018{\natexlab{a}})Haarnoja, Zhou, Abbeel, and
  Levine]{sac}
Haarnoja, T., Zhou, A., Abbeel, P., and Levine, S.
\newblock Soft actor-critic: Off-policy maximum entropy deep reinforcement
  learning with a stochastic actor.
\newblock \emph{arXiv preprint arXiv:1801.01290}, 2018{\natexlab{a}}.

\bibitem[Haarnoja et~al.(2018{\natexlab{b}})Haarnoja, Zhou, Hartikainen,
  Tucker, Ha, Tan, Kumar, Zhu, Gupta, Abbeel, et~al.]{sac-alg}
Haarnoja, T., Zhou, A., Hartikainen, K., Tucker, G., Ha, S., Tan, J., Kumar,
  V., Zhu, H., Gupta, A., Abbeel, P., et~al.
\newblock Soft actor-critic algorithms and applications.
\newblock \emph{arXiv preprint arXiv:1812.05905}, 2018{\natexlab{b}}.

\bibitem[Hausman et~al.(2018)Hausman, Springenberg, Wang, Heess, and
  Riedmiller]{hierarchical-subgoals-robot}
Hausman, K., Springenberg, J.~T., Wang, Z., Heess, N., and Riedmiller, M.
\newblock Learning an embedding space for transferable robot skills.
\newblock In \emph{International Conference on Learning Representations}, 2018.

\bibitem[Hessel et~al.(2018)Hessel, Modayil, Van~Hasselt, Schaul, Ostrovski,
  Dabney, Horgan, Piot, Azar, and Silver]{rainbow}
Hessel, M., Modayil, J., Van~Hasselt, H., Schaul, T., Ostrovski, G., Dabney,
  W., Horgan, D., Piot, B., Azar, M., and Silver, D.
\newblock Rainbow: Combining improvements in deep reinforcement learning.
\newblock In \emph{Proceedings of the AAAI Conference on Artificial
  Intelligence}, volume~32, 2018.

\bibitem[Kulkarni et~al.(2016)Kulkarni, Narasimhan, Saeedi, and
  Tenenbaum]{hierarchical-subgoals-intrinsic-1}
Kulkarni, T.~D., Narasimhan, K., Saeedi, A., and Tenenbaum, J.
\newblock Hierarchical deep reinforcement learning: Integrating temporal
  abstraction and intrinsic motivation.
\newblock \emph{Advances in neural information processing systems},
  29:\penalty0 3675--3683, 2016.

\bibitem[Lakshminarayanan et~al.(2017)Lakshminarayanan, Sharma, and
  Ravindran]{action-reps-discrete}
Lakshminarayanan, A., Sharma, S., and Ravindran, B.
\newblock Dynamic action repetition for deep reinforcement learning.
\newblock In \emph{Proceedings of the AAAI Conference on Artificial
  Intelligence}, volume~31, 2017.

\bibitem[LeCun et~al.(2015)LeCun, Bengio, and Hinton]{lecun-deep}
LeCun, Y., Bengio, Y., and Hinton, G.
\newblock Deep learning.
\newblock \emph{nature}, 521\penalty0 (7553):\penalty0 436--444, 2015.

\bibitem[Lee et~al.(2020)Lee, Lee, and Kim]{actpers1}
Lee, J., Lee, B.-J., and Kim, K.-E.
\newblock Reinforcement learning for control with multiple frequencies.
\newblock In \emph{Thirty-fourth Conference on Neural Information Processing
  Systems (NeurIPS 2020)}. Neural information processing systems foundation,
  2020.

\bibitem[Lillicrap et~al.(2015)Lillicrap, Hunt, Pritzel, Heess, Erez, Tassa,
  Silver, and Wierstra]{ddpg}
Lillicrap, T.~P., Hunt, J.~J., Pritzel, A., Heess, N., Erez, T., Tassa, Y.,
  Silver, D., and Wierstra, D.
\newblock Continuous control with deep reinforcement learning.
\newblock \emph{arXiv preprint arXiv:1509.02971}, 2015.

\bibitem[Mahmood et~al.(2018)Mahmood, Korenkevych, Vasan, Ma, and
  Bergstra]{rl-real-world-design}
Mahmood, A.~R., Korenkevych, D., Vasan, G., Ma, W., and Bergstra, J.
\newblock Benchmarking reinforcement learning algorithms on real-world robots.
\newblock \emph{arXiv preprint arXiv:1809.07731}, 2018.

\bibitem[Metelli et~al.(2020)Metelli, Mazzolini, Bisi, Sabbioni, and
  Restelli]{actpers0}
Metelli, A.~M., Mazzolini, F., Bisi, L., Sabbioni, L., and Restelli, M.
\newblock Control frequency adaptation via action persistence in batch
  reinforcement learning.
\newblock In \emph{International Conference on Machine Learning}, pp.\
  6862--6873. PMLR, 2020.

\bibitem[Mnih et~al.(2013)Mnih, Kavukcuoglu, Silver, Graves, Antonoglou,
  Wierstra, and Riedmiller]{dqn}
Mnih, V., Kavukcuoglu, K., Silver, D., Graves, A., Antonoglou, I., Wierstra,
  D., and Riedmiller, M.
\newblock Playing atari with deep reinforcement learning.
\newblock \emph{arXiv preprint arXiv:1312.5602}, 2013.

\bibitem[Mnih et~al.(2016)Mnih, Badia, Mirza, Graves, Lillicrap, Harley,
  Silver, and Kavukcuoglu]{A3c}
Mnih, V., Badia, A.~P., Mirza, M., Graves, A., Lillicrap, T., Harley, T.,
  Silver, D., and Kavukcuoglu, K.
\newblock Asynchronous methods for deep reinforcement learning.
\newblock In \emph{International conference on machine learning}, pp.\
  1928--1937, 2016.

\bibitem[Nachum et~al.(2018)Nachum, Gu, Lee, and Levine]{HIRO}
Nachum, O., Gu, S.~S., Lee, H., and Levine, S.
\newblock Data-efficient hierarchical reinforcement learning.
\newblock In \emph{Advances in neural information processing systems}, pp.\
  3303--3313, 2018.

\bibitem[Neunert et~al.(2020)Neunert, Abdolmaleki, Wulfmeier, Lampe,
  Springenberg, Hafner, Romano, Buchli, Heess, and Riedmiller]{actreps_mot2}
Neunert, M., Abdolmaleki, A., Wulfmeier, M., Lampe, T., Springenberg, T.,
  Hafner, R., Romano, F., Buchli, J., Heess, N., and Riedmiller, M.
\newblock Continuous-discrete reinforcement learning for hybrid control in
  robotics.
\newblock In \emph{Conference on Robot Learning}, pp.\  735--751. PMLR, 2020.

\bibitem[Precup(2001)]{opt-intro2}
Precup, D.
\newblock Temporal abstraction in reinforcement learning.
\newblock 2001.

\bibitem[Precup et~al.(1997)Precup, Sutton, and Singh]{macro-intro}
Precup, D., Sutton, R.~S., and Singh, S.~P.
\newblock Planning with closed-loop macro actions.
\newblock In \emph{Working notes of the 1997 AAAI Fall Symposium on
  Model-directed Autonomous Systems}, pp.\  70--76, 1997.

\bibitem[Schoknecht \& Riedmiller(2002)Schoknecht and Riedmiller]{actreps_mot0}
Schoknecht, R. and Riedmiller, M.
\newblock Speeding-up reinforcement learning with multi-step actions.
\newblock In \emph{International Conference on Artificial Neural Networks},
  pp.\  813--818. Springer, 2002.

\bibitem[Schoknecht \& Riedmiller(2003)Schoknecht and Riedmiller]{actreps_mot1}
Schoknecht, R. and Riedmiller, M.
\newblock Reinforcement learning on explicitly specified time scales.
\newblock \emph{Neural Computing \& Applications}, 12\penalty0 (2):\penalty0
  61--80, 2003.

\bibitem[Sharma et~al.(2017)Sharma, Lakshminarayanan, and
  Ravindran]{action-reps-policy-factor}
Sharma, S., Lakshminarayanan, A.~S., and Ravindran, B.
\newblock Learning to repeat: Fine grained action repetition for deep
  reinforcement learning.
\newblock \emph{arXiv preprint arXiv:1702.06054}, 2017.

\bibitem[Silver et~al.(2014)Silver, Lever, Heess, Degris, Wierstra, and
  Riedmiller]{dpg}
Silver, D., Lever, G., Heess, N., Degris, T., Wierstra, D., and Riedmiller, M.
\newblock Deterministic policy gradient algorithms.
\newblock 2014.

\bibitem[Silver et~al.(2017)Silver, Hubert, Schrittwieser, Antonoglou, Lai,
  Guez, Lanctot, Sifre, Kumaran, Graepel, et~al.]{alphazero}
Silver, D., Hubert, T., Schrittwieser, J., Antonoglou, I., Lai, M., Guez, A.,
  Lanctot, M., Sifre, L., Kumaran, D., Graepel, T., et~al.
\newblock Mastering chess and shogi by self-play with a general reinforcement
  learning algorithm.
\newblock \emph{arXiv preprint arXiv:1712.01815}, 2017.

\bibitem[Sutton \& Barto(2018)Sutton and Barto]{sutton-reinforcement-n-step}
Sutton, R.~S. and Barto, A.~G.
\newblock \emph{Reinforcement learning: An introduction}.
\newblock MIT press, 2018.

\bibitem[Sutton et~al.(1999)Sutton, Precup, and Singh]{opt-intro1}
Sutton, R.~S., Precup, D., and Singh, S.
\newblock Between mdps and semi-mdps: A framework for temporal abstraction in
  reinforcement learning.
\newblock \emph{Artificial intelligence}, 112\penalty0 (1-2):\penalty0
  181--211, 1999.

\bibitem[Sutton et~al.(2000)Sutton, McAllester, Singh, and Mansour]{pg-thm}
Sutton, R.~S., McAllester, D.~A., Singh, S.~P., and Mansour, Y.
\newblock Policy gradient methods for reinforcement learning with function
  approximation.
\newblock In \emph{Advances in neural information processing systems}, pp.\
  1057--1063, 2000.

\bibitem[Sweatt(2009)]{human-motor-learning}
Sweatt, J.~D.
\newblock \emph{Mechanisms of memory}.
\newblock Academic Press, 2009.

\bibitem[Tassa et~al.(2018)Tassa, Doron, Muldal, Erez, Li, Casas, Budden,
  Abdolmaleki, Merel, Lefrancq, et~al.]{dmc}
Tassa, Y., Doron, Y., Muldal, A., Erez, T., Li, Y., Casas, D. d.~L., Budden,
  D., Abdolmaleki, A., Merel, J., Lefrancq, A., et~al.
\newblock Deepmind control suite.
\newblock \emph{arXiv preprint arXiv:1801.00690}, 2018.

\bibitem[Tennenholtz \& Mannor(2019)Tennenholtz and Mannor]{demos-act-rep}
Tennenholtz, G. and Mannor, S.
\newblock The natural language of actions.
\newblock \emph{arXiv preprint arXiv:1902.01119}, 2019.

\bibitem[Tessler et~al.(2017)Tessler, Givony, Zahavy, Mankowitz, and
  Mannor]{hierarchical-subgoals-mine}
Tessler, C., Givony, S., Zahavy, T., Mankowitz, D., and Mannor, S.
\newblock A deep hierarchical approach to lifelong learning in minecraft.
\newblock In \emph{Proceedings of the AAAI Conference on Artificial
  Intelligence}, volume~31, 2017.

\bibitem[Vezhnevets et~al.(2016)Vezhnevets, Mnih, Osindero, Graves, Vinyals,
  Agapiou, et~al.]{macro-STRAWS}
Vezhnevets, A., Mnih, V., Osindero, S., Graves, A., Vinyals, O., Agapiou, J.,
  et~al.
\newblock Strategic attentive writer for learning macro-actions.
\newblock \emph{Advances in neural information processing systems},
  29:\penalty0 3486--3494, 2016.

\bibitem[Vezhnevets et~al.(2017)Vezhnevets, Osindero, Schaul, Heess, Jaderberg,
  Silver, and Kavukcuoglu]{FuN}
Vezhnevets, A.~S., Osindero, S., Schaul, T., Heess, N., Jaderberg, M., Silver,
  D., and Kavukcuoglu, K.
\newblock Feudal networks for hierarchical reinforcement learning.
\newblock \emph{arXiv preprint arXiv:1703.01161}, 2017.

\bibitem[Whitney et~al.(2019)Whitney, Agarwal, Cho, and
  Gupta]{actseq-rep-dynamics-aware-emb}
Whitney, W., Agarwal, R., Cho, K., and Gupta, A.
\newblock Dynamics-aware embeddings.
\newblock \emph{arXiv preprint arXiv:1908.09357}, 2019.

\bibitem[Ziebart(2010)]{maxentobj}
Ziebart, B.~D.
\newblock Modeling purposeful adaptive behavior with the principle of maximum
  causal entropy.
\newblock 2010.

\end{thebibliography}
